\documentclass[lettersize,journal]{IEEEtran}
\usepackage{amsmath,amsfonts}
\usepackage{algorithmic}
\usepackage{algorithm}
\usepackage{array}
\usepackage{amssymb}
\usepackage{pifont}
\usepackage[caption=false,font=normalsize,labelfont=sf,textfont=sf]{subfig}
\usepackage{textcomp}
\usepackage{stfloats}
\usepackage{url}
\usepackage{verbatim}
\usepackage{graphicx}
\usepackage{cite}
\usepackage{booktabs}
\usepackage{multirow}
\usepackage{color}
\usepackage{makecell}

\usepackage{adjustbox}
\usepackage{xcolor}
\usepackage{enumitem}

\usepackage[switch]{lineno}

\usepackage[
            colorlinks,
            linkcolor=blue,
            anchorcolor=blue,
            citecolor=blue,
            urlcolor=blue
            ]{hyperref}
\newenvironment{majorrevision}{\begingroup\color{black}}{\endgroup}

\begin{document}

\title{A Systematic Survey on Large Language Models for Evolutionary Optimization: From Modeling to Solving}

\author{
Yisong Zhang, Ran Cheng, Guoxing Yi, and Kay Chen Tan

}

\maketitle

\begin{abstract}
Large language models (LLMs) are increasingly integrated with evolutionary computation to support optimization tasks\footnote{This survey primarily focuses on \textit{evolutionary optimization}, i.e., optimization based on evolutionary computation. For brevity, we use the term \textit{optimization} throughout to denote this scope.}.
However, existing surveys typically examine isolated roles of LLMs and do not provide a unified view that connects optimization modeling with optimization solving.
To address this gap, we systematically review recent developments through a workflow-oriented framework.
First, we organize the literature into two primary stages: LLMs for optimization modeling and LLMs for optimization solving\footnote{In this survey, the terms \textit{optimization modeling} and \textit{optimization solving} are used as concise forms of \textit{optimization problem modeling} and \textit{optimization problem solving}, respectively.}.
Second, we divide the solving stage into three paradigms according to the role of the LLM: stand-alone optimizers, low-level components embedded in optimization algorithms, and high-level managers for algorithm selection and generation.
Third, we analyze representative methods, identify their technical limitations, and clarify their relationships with traditional optimization approaches.
We further substantiate this taxonomy through benchmark systematization, baseline comparisons, and practitioner-oriented guidance, and we review interdisciplinary applications across the natural sciences, engineering, and machine learning.
Based on the resulting analysis, we identify research directions toward dynamic, self-evolving, and agentic optimization ecosystems.
An up-to-date collection of related literature is maintained at \url{https://github.com/ishmael233/LLM4OPT}.
\end{abstract}

\begin{IEEEkeywords}
Large language models, evolutionary algorithms, optimization modeling, optimization solving.
\end{IEEEkeywords}

\section{Introduction}

Optimization underpins complex decision-making across engineering design \cite{chai2013evolutionary}, economic planning \cite{mahor2009economic}, scientific discovery \cite{terayama2021black}, and many other domains \cite{chen2022evolutionary, cambria2014computational}.
A practical optimization workflow first abstracts a real-world problem into a mathematical model and then applies an optimization algorithm to solve that model.
These algorithms can be broadly divided into exact and approximate methods \cite{festa2014brief}.
Exact methods can provide global-optimality guarantees, but their computational cost often grows prohibitively with problem scale.
Approximate methods, including heuristics \cite{festa2014brief}, can address larger and more complex problems efficiently, but they generally do not guarantee convergence to a global optimum.
Moreover, the No Free Lunch theorem \cite{wolpert2002no} implies that no single algorithm performs best across all problem classes.
Consequently, effective deployment often requires substantial expertise in algorithm configuration \cite{hutter2009paramils}, selection \cite{kerschke2019automated}, and customized design \cite{muller1981heuristics}.
This dependence on expert knowledge remains a major barrier to transferring optimization methods from research to practice.

Machine learning (ML) has been introduced into evolutionary computation to reduce this dependence on expert knowledge \cite{zhang2011evolutionary, ma2025toward}.
At the component level, ML supports population initialization \cite{mundhenk2021symbolic}, fitness evaluation \cite{wang2022surrogate}, operator selection \cite{pettinger2002controlling}, parameter control \cite{eiben2006reinforcement}, and solution manipulation \cite{chen2017learning}.
At the strategy level, it enables automated algorithm selection \cite{lagoudakis2000algorithm} and algorithm generation \cite{chen2024symbol}.
Most existing approaches rely on reinforcement learning \cite{mazyavkina2021reinforcement} or supervised learning \cite{li2024pretrained, han2026enhancing} and are trained on narrowly defined problem distributions.
As a result, their performance often degrades on unseen problems, while adaptation requires costly and time-consuming retraining.
Thus, conventional learning-based methods reduce some forms of manual design but do not fully resolve the generalization challenge.

Large language models (LLMs) offer a complementary route because they combine broad pretrained knowledge with semantic understanding, compositional abstraction, and in-context learning \cite{zhao2023survey}.
These capabilities support zero-shot or few-shot transfer across heterogeneous tasks \cite{patil2024review}.
They are particularly relevant when objectives, constraints, or design criteria are expressed in natural language or structured symbols, and when the workflow requires readable candidate generation, explanation, or cross-domain priors, as in drug discovery \cite{mi2022pangu} and finance \cite{wu2023bloomberggpt}.

Within optimization, LLMs can contribute to both modeling and solving.
For modeling, they translate informal problem descriptions into formal mathematical formulations \cite{huang2025orlm}.
For solving, they can operate as stand-alone iterative optimizers for language-structured tasks \cite{yang2023large} or as components that augment algorithmic search \cite{liu2024evolution}.
However, this integration remains at an early stage and exhibits clear limitations.
Stand-alone LLM optimizers often underperform classical algorithms \cite{huang2024exploring}, while domain-specific fine-tuning depends on datasets that are costly to construct and validate \cite{huang2025orlm}.
These limitations indicate the need for a unified analysis of where LLMs are effective, how they should interact with traditional methods, and which research gaps remain.

\begin{figure*}[ht!]
    \centering
    \includegraphics[width=0.93\textwidth]{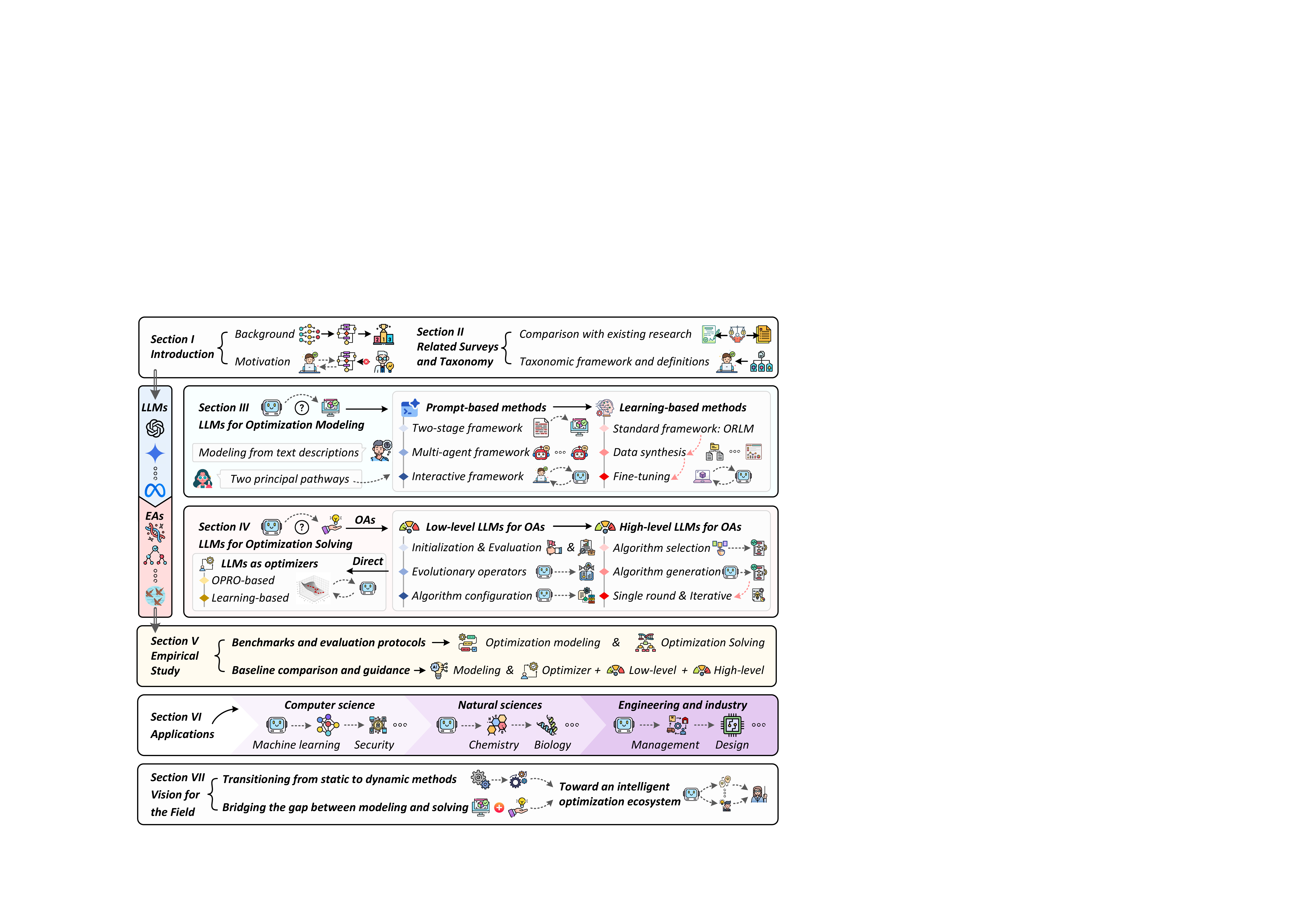}
    \caption{Overall organization of this survey.}
    \label{fig:intro}
\end{figure*}

To provide this analysis, we integrate a systematic literature review with benchmark systematization, controlled baseline comparisons, and practitioner-oriented guidance.
The main contributions are summarized as follows:
\begin{itemize}
\item We adopt a modeling-to-solving perspective that covers the complete optimization workflow. Based on this perspective, we divide the field into optimization modeling and optimization solving, and further organize the solving stage into LLMs as optimizers, low-level LLM-assisted algorithms, and high-level LLM-assisted algorithm selection and generation.
We ground the taxonomy empirically through data collection and controlled experiments across the modeling-to-solving spectrum. These analyses expose concrete failure modes, including limited numerical precision, weak dimensional scalability, and feasibility constraints.
We translate the resulting evidence into practitioner-oriented method-selection guidance and identify research directions toward dynamic, self-evolving, and agentic optimization ecosystems.
\end{itemize}

Fig.~\ref{fig:intro} illustrates the organization of this survey.
Section~\ref{Related Surveys and Taxonomy} reviews related surveys and defines the proposed taxonomy.
Sections~\ref{LLMs for Optimization Modeling} and~\ref{LLMs for Optimization Solving} examine LLMs for optimization modeling and solving, respectively.
Section~\ref{Empirical Study and Guidance} systematizes benchmarks, compares representative baselines, and derives practitioner-oriented guidance.
Section~\ref{Applications} reviews interdisciplinary applications.
Finally, Section~\ref{Vision for the Field} discusses future research directions, and Section~\ref{Conclusion} summarizes the main findings.

\section{Related Surveys and Taxonomy}
\label{Related Surveys and Taxonomy}

\begin{majorrevision}
\subsection{Survey Scope and Methodology}
\label{Survey Scope and Methodology}

We conducted a structured literature search to obtain broad but technically focused coverage of the field.
The search covered IEEE Xplore, the ACM Digital Library, SpringerLink, ScienceDirect, Google Scholar, and arXiv.
We combined model-side keywords, including \emph{large language model}, \emph{LLM}, and \emph{foundation model}, with task-side keywords, including \emph{optimization}, \emph{evolutionary algorithm/computation}, \emph{metaheuristic}, \emph{heuristic design}, \emph{optimization modeling}, and \emph{algorithm selection/generation}.
The search period extended from January 2023 to September 2025, which covers the rapid expansion of the field following advances in prompting and in-context learning.

\begin{figure}[t]
\begin{center}
\includegraphics[width=0.45\textwidth]{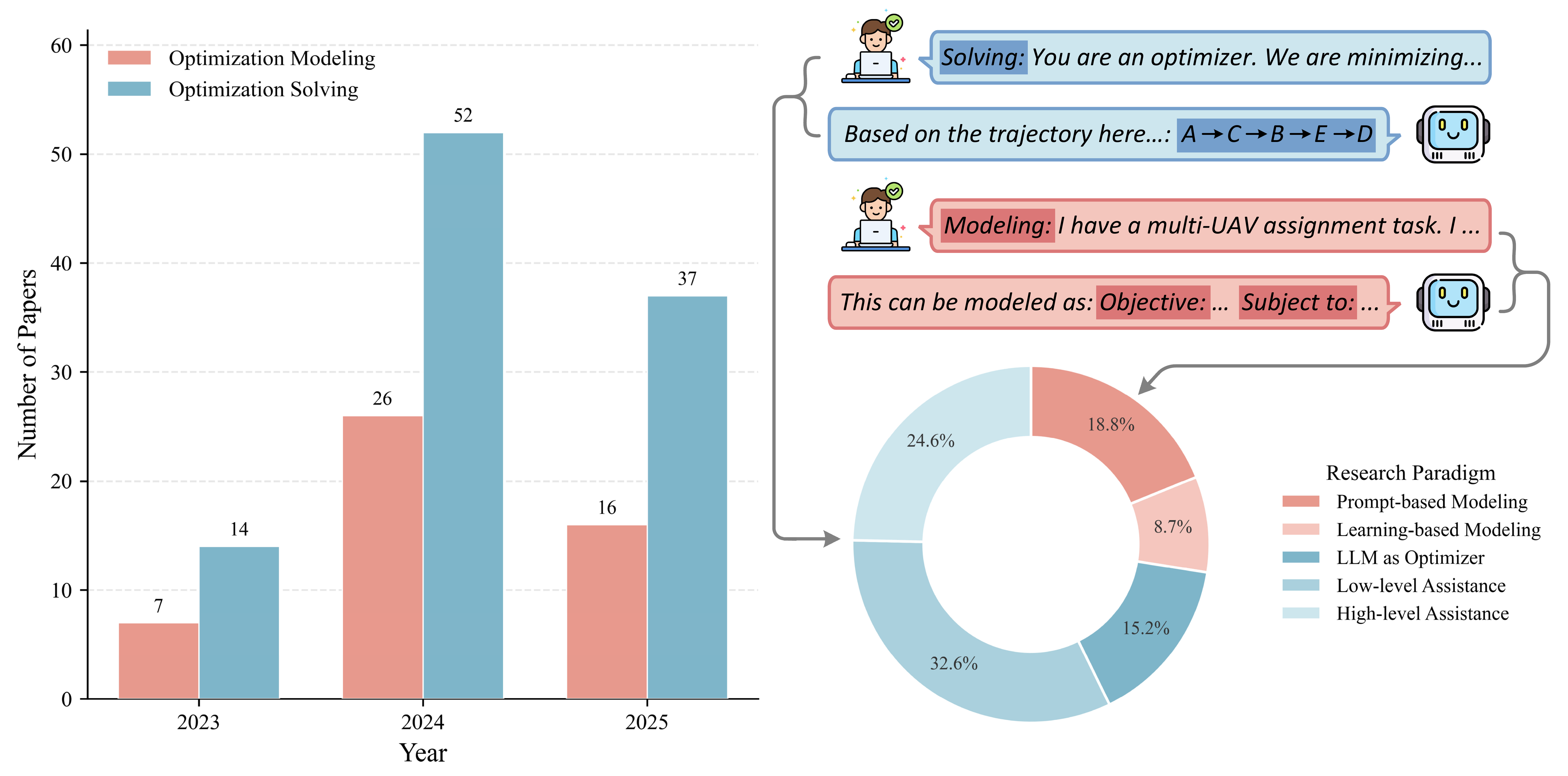}
\end{center}
\caption{Publication trends and paradigm distribution in the surveyed literature.}
\label{fig:bar_pie}
\end{figure}

We applied two inclusion criteria.
First, a study had to use one or more LLMs as an integral component of optimization modeling or solving.
Second, it had to report a methodological contribution or an empirical evaluation.
We excluded studies that mentioned LLMs without assigning them a concrete optimization role, as well as short abstracts and non-peer-reviewed notes that lacked sufficient methodological detail.
After deduplication and manual screening of titles, abstracts, and full texts, we retained 152 representative papers as the corpus of this survey.

Because the literature is expanding rapidly, no survey can guarantee exhaustive coverage of every related study.
We therefore aim to characterize the main technical landscape through a representative corpus of methodologically substantive work.
Fig.~\ref{fig:bar_pie} summarizes the publication timeline and the distribution of papers across the proposed taxonomy.
The results show that most studies in the corpus were published during the final two years of the search period, which indicates a sharp recent increase in research on the integration of LLMs with optimization.
\end{majorrevision}

\subsection{Related Surveys and Our Perspective}
\label{Related Surveys and Our Perspective}

Several surveys have examined LLMs for optimization, as summarized in Table~\ref{tab:related}.
Wu \textit{et al.} \cite{wu2025evolutionary} and Huang \textit{et al.} \cite{huang2024large} adopt a bidirectional perspective but focus, on the optimization side, on LLMs as solvers or algorithm generators.
Yu \textit{et al.} \cite{yu2024deep} use a similar classification, whereas Chao \textit{et al.} \cite{chao2024large} focus on evolutionary algorithms (EAs) and emphasize LLMs as operators.
Liu \textit{et al.} \cite{liu2024systematic} broadly review LLMs in algorithm design, and Ma \textit{et al.} \cite{ma2025toward} survey MetaBBO with an emphasis on reinforcement learning.

\begin{majorrevision}
Two structural gaps remain across these reviews.
First, most studies focus on optimization solving, while optimization modeling receives limited attention or is treated as a peripheral application.
Second, reviews of the solving stage commonly emphasize one role of the LLM without placing stand-alone optimization, component-level assistance, and high-level orchestration within a common framework.
Consequently, the relationships among these complementary roles remain insufficiently characterized.
\end{majorrevision}

\begin{table}[t!]
    \centering
    \caption{Comparison of related surveys on LLMs for optimization. This work systematically covers four categories: LLMs for Modeling (LM), LLMs as Optimizers (LO), Low-level LLMs for Optimization Algorithms (LL), and High-level LLMs for Optimization Algorithms (HL).}
    \label{tab:related}
    \renewcommand{\arraystretch}{1.2}
    \begin{adjustbox}{max width=\linewidth}
    \begin{tabular}{c|c|c|c|c|c}
        \toprule
        \textbf{Ref.} & \textbf{Venue} & \textbf{LM} & \textbf{LO} & \textbf{LL} & \textbf{HL} \\
        \midrule
        Wu \textit{et al.} \cite{wu2025evolutionary} & TEVC, 2024 & {} & \ding{51} & {} & \ding{51} \\
        Huang \textit{et al.} \cite{huang2024large} & SWEVO, 2024 & {} & \ding{51} & {} & \ding{51} \\
        Chao \textit{et al.} \cite{chao2024large} & Research, 2024 & {} & {} & \ding{51} & {} \\
        Yu \textit{et al.} \cite{yu2024deep} & arXiv, 2024 & {} & \ding{51} & \ding{51} & \ding{51} \\
        Liu \textit{et al.} \cite{liu2024systematic} & arXiv, 2024 & {} & \ding{51} & \ding{51} & \ding{51} \\
        Ma \textit{et al.} \cite{ma2025toward} & TEVC, 2024 & {} & \ding{51} & {} & \ding{51} \\
        \midrule
        Our work & arXiv, 2025 & \ding{51} & \ding{51} & \ding{51} & \ding{51} \\
        \bottomrule
    \end{tabular}
    \end{adjustbox}
\end{table}

\begin{figure}[t]
\begin{center}
\includegraphics[width=0.45\textwidth]{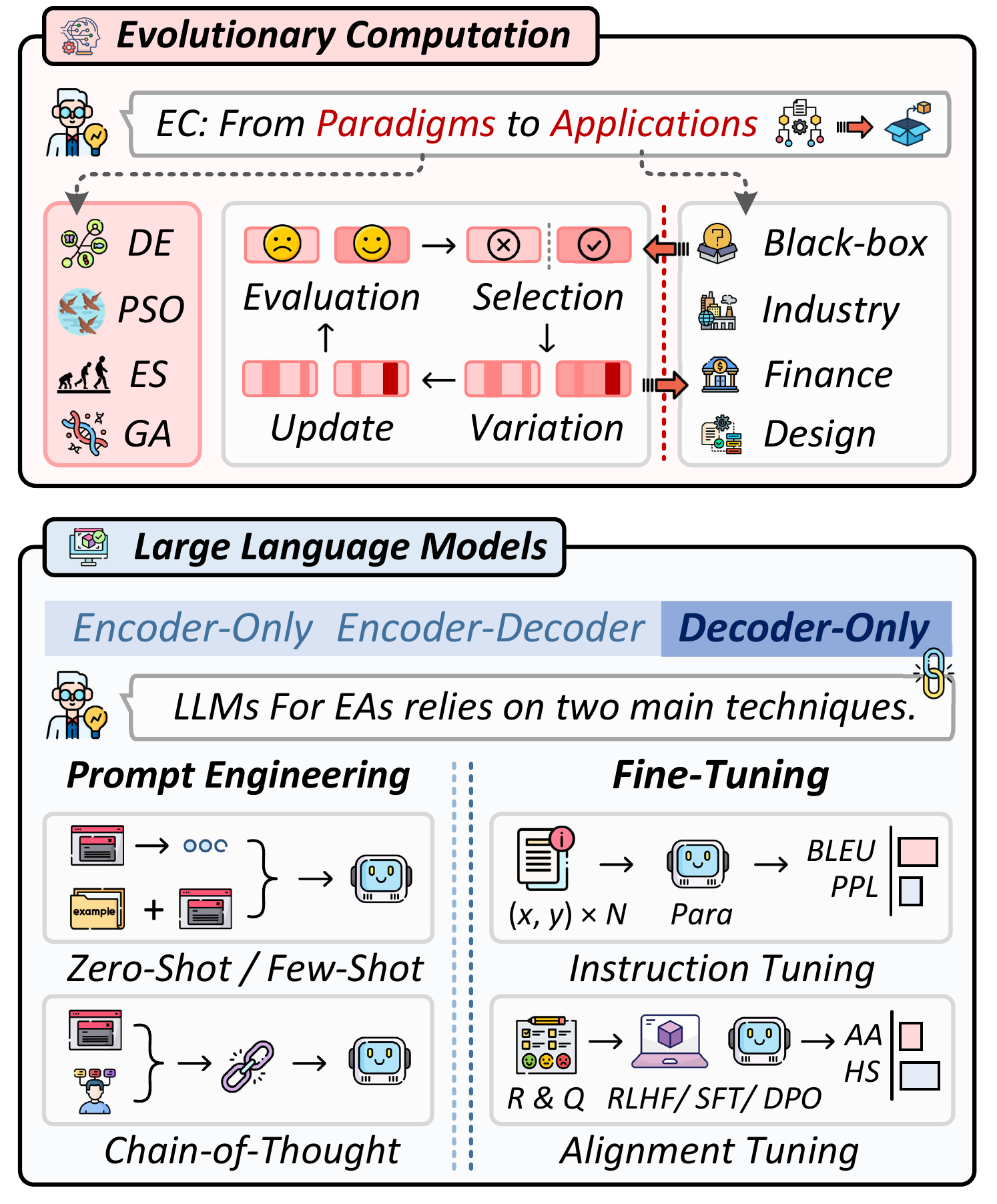}
\end{center}
\caption{Technological dependencies of LLMs for optimization, including EA paradigm and workflow, LLM architecture, and related enabling technologies.}
\label{fig:background_v2}
\end{figure}

To address these gaps, this survey adopts a workflow-oriented perspective that follows the full optimization lifecycle from natural-language problem description to mathematical modeling and algorithmic solving.
This perspective extends prior reviews in two respects.
First, it treats optimization modeling as a primary stage rather than a peripheral application.
Second, it organizes optimization solving as a spectrum of three roles: LLMs as optimizers, low-level LLM assistance, and high-level LLM assistance.
\begin{majorrevision}
This common frame makes the roles directly comparable and reveals an important missing link: current methods rarely form an end-to-end loop that connects automated modeling with LLM-assisted solving.
\end{majorrevision}
Section~\ref{Vision for the Field} revisits this gap.
\begin{majorrevision}
Fig.~\ref{fig:background_v2} summarizes the supporting technologies, including the EA workflow, LLM architectures, and related enabling methods, while Supplementary Document I provides further background on EAs and LLMs.
Based on the same workflow-oriented perspective, Section~\ref{Empirical Study and Guidance} evaluates representative methods and derives practical method-selection guidance.
\end{majorrevision}

\subsection{Taxonomy and Definitions}
\label{Taxonomy and Definitions}

Based on this workflow-oriented perspective, we divide the field into two top-level categories: LLMs for optimization modeling and LLMs for optimization solving.
We further divide the solving stage into three paradigms according to the role of the LLM.
The four categories are defined as follows:
\begin{itemize}
    \item \textbf{LLMs for Optimization Modeling} transform unstructured natural-language problem descriptions into mathematical optimization models that machines can interpret and solve.
    This category addresses the transition from ambiguous language to precise variables, objectives, and constraints.
    Existing methods are primarily prompt-based or learning-based.
    Prompt-based methods use designed instructions, structured workflows, or multiple agents and can be deployed without additional training.
    Learning-based methods use data synthesis and fine-tuning to improve reliability, but they introduce training and maintenance costs.

    \item \textbf{LLMs as Optimizers} use LLMs as stand-alone optimizers that iteratively generate candidate solutions through natural-language interaction, without embedding them in a traditional optimization framework.
    These methods exploit in-context learning and reasoning over previous solutions and feedback.
    Their direct formulation makes them simple to deploy, but their performance depends strongly on the underlying model and often scales poorly with numerical dimensionality.

    \item \textbf{Low-level LLM-assisted Optimization Algorithms} embed LLMs within established optimization algorithms to support specific components.
    Typical roles include population initialization, evolutionary variation, parameter control, algorithm configuration, and fitness evaluation.
    In this paradigm, the optimization algorithm retains control of the overall search, while the LLM contributes semantic priors, domain knowledge, or adaptive local decisions.

    \item \textbf{High-level LLM-assisted Optimization Algorithms} use LLMs for orchestration or design at the algorithm level.
    This category includes algorithm selection, in which an LLM chooses a suitable method from a portfolio, and algorithm generation, in which an LLM constructs or refines an optimization algorithm for a target task.
    The LLM therefore acts as a selector or designer rather than as an internal search component.
\end{itemize}

\section{LLMs for Optimization Modeling}
\label{LLMs for Optimization Modeling}

\begin{figure*}[ht!]
    \centering
    \includegraphics[width=0.90\textwidth]{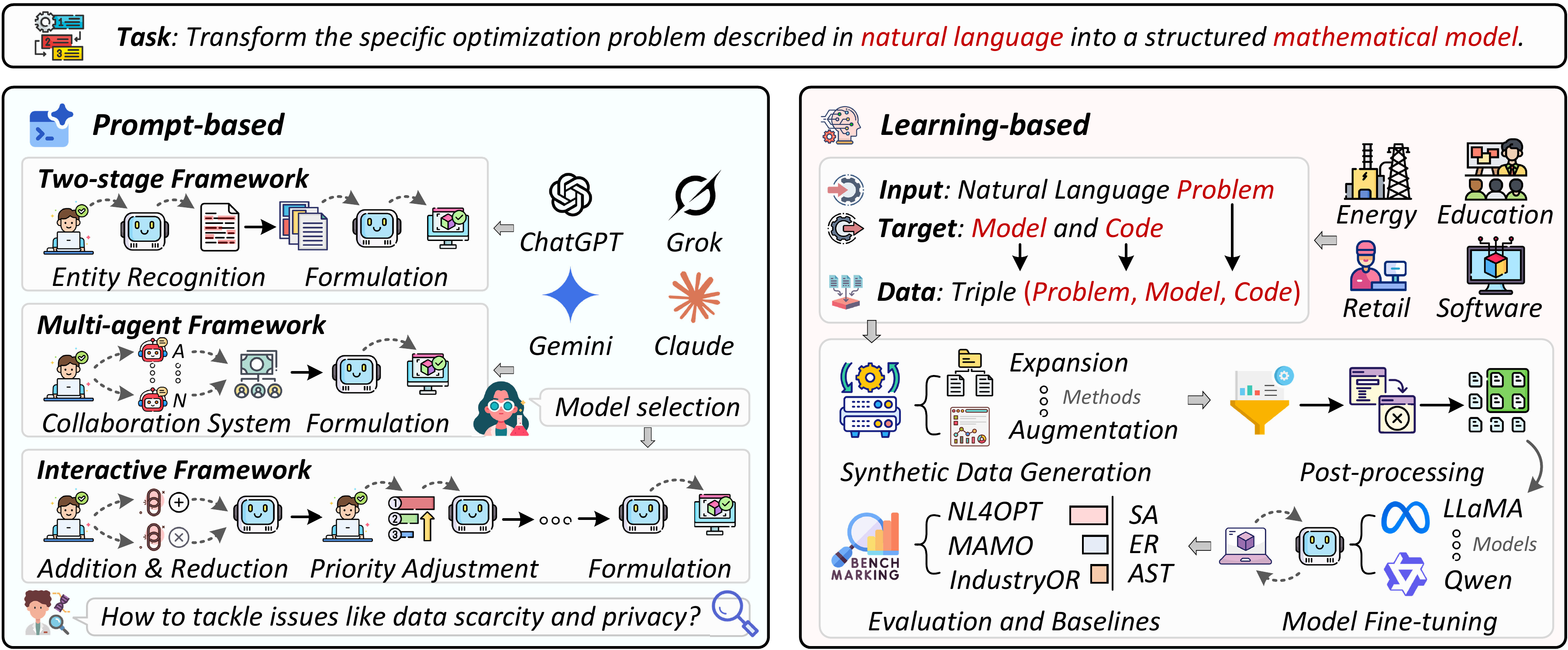}
    \caption{Illustration of LLMs for optimization modeling. Approaches are divided into two categories: (i) prompt-based methods, which are typically implemented via two-stage prompting, multi-agent collaboration, or interactive frameworks; and (ii) learning-based methods, which generally follow a workflow involving data synthesis, model fine-tuning, and evaluation.}
    \label{fig:formulation}
\end{figure*}

Optimization modeling is the first computational stage of the optimization workflow, but it traditionally requires substantial domain and mathematical expertise.
LLMs provide a mechanism for automating parts of this process by mapping natural-language specifications to formal variables, objectives, and constraints.
This section reviews two main paradigms.
Prompt-based methods, discussed in Section~\ref{Prompt-based Methods}, guide LLMs through designed instructions, two-stage workflows, multi-agent collaboration, or interactive refinement.
Learning-based methods, discussed in Section~\ref{Learning-based Methods}, fine-tune LLMs to generate mathematical formulations directly and commonly rely on synthetic training data.
Section~\ref{Challenges} then compares the limitations and deployment trade-offs of the two paradigms.
Fig.~\ref{fig:formulation} illustrates their main workflows, and Supplementary Document II summarizes representative studies.

\subsection{Prompt-based Methods}
\label{Prompt-based Methods}

Early work on automated optimization modeling attempted to reduce expert involvement by inferring problem structure from feasible solutions or by using EAs.
Pawlak \textit{et al.} studied the inference of optimization models from solution samples \cite{pawlak2017automatic} and the generation of constraints through grammatical evolution \cite{pawlak2021grammatical}.
These studies established the feasibility of partial automation, but their performance depends strongly on sample quality and coverage \cite{li2023synthesizing}.
Moreover, they primarily generate constraints rather than translate natural-language descriptions into complete mathematical models.

Subsequent research shifted toward the direct translation of unstructured language into mathematical optimization models.
Ramamonjison \textit{et al.} proposed OptGen \cite{ramamonjison2022augmenting} and initiated the NL4OPT competition \cite{ramamonjison2023nl4opt}, which decomposed modeling into named-entity recognition (NER) followed by model generation from annotated descriptions.
Early systems used smaller language models, including BERT \cite{wang2023opd, doan2022vtcc} and BART \cite{gangwar2023highlighting, ning2023novel}.
Almonacid \textit{et al.} \cite{almonacid2023towards} later introduced a single-step LLM approach that prompts GPT-3.5 to generate a model and uses a solver to verify the result.
Although this approach is effective on simple instances, a single generation step provides insufficient accuracy and robustness for more complex formulations.

To improve reliability, researchers developed two-stage frameworks that combine LLMs with specialized extraction modules.
Holy Grail 2.0 \cite{tsouros2023holy} decomposes the workflow into entity-relation identification, problem formalization, and code generation.
Li \textit{et al.} \cite{li2023synthesizing} extended this structure to mixed-integer linear programming by fine-tuning a model to classify constraints, including logical constraints and binary variables that earlier methods often omitted.
Similar two-stage workflows have been applied to energy management \cite{jin2024democratizing} and travel planning \cite{de2024trip}.

Two further extensions relax the fixed pipeline.
Multi-agent frameworks assign complementary modeling roles to several LLM agents and use cross-checking to improve coverage and consistency.
Interactive frameworks instead maintain a dialogue with the user so that the formulation can be revised as requirements and preferences change.

Multi-agent modeling commonly separates formulation, implementation, and verification across specialized agents \cite{liang2025llm}.
The Chain-of-Experts (CoE) framework \cite{xiao2023chain} coordinates 11 expert types through a conductor agent, constructs a forward chain of thought, and applies backward reflection to correct errors and inconsistencies.
OptiMUS \cite{ahmaditeshnizi2024optimus} and OptiMUS-0.3 \cite{ahmaditeshnizi2024optimus03} similarly use a conductor to coordinate modeling, programming, and evaluation.
However, conductor-driven interaction can produce variable workflows and insufficient mathematical precision.
To improve process control, ORMind \cite{wang2025ormind} uses a structured, cognitively inspired workflow with counterfactual reasoning.
Other studies focus specifically on verification: Mostajabdaveh \textit{et al.} \cite{mostajabdaveh2024optimization} replace solver-based checking with peer verification among agents, while Talebi \textit{et al.} \cite{talebi2025large} use independent reviewer agents to evaluate stochastic optimization models.

Interactive modeling addresses problems for which the initial specification is incomplete or changes during use.
This requirement is common in conference scheduling, travel planning, and other preference-sensitive tasks.
OptLLM \cite{zhang2024solving} supports both single-shot and interactive inputs through a refinement module that progressively clarifies the problem description, followed by converter and response modules for modeling and verification.
MeetMeta \cite{lawless2024want} requires the LLM to select among five actions after each user message, return a solution, explain unmet preferences, and accept subsequent modifications.
Related work translates user priorities into constraints during dialogue, which allows users to examine trade-offs in applications such as ridesharing coordination \cite{gomez2025values}.

Recent prompt-based research has also targeted generation and verification as separate technical bottlenecks \cite{deng2024cafa}.
For generation, Astorga \textit{et al.} \cite{astorga2024autoformulation} combine LLMs with Monte Carlo Tree Search to explore a hierarchical hypothesis space through symbolic pruning and LLM-based evaluation.
Li \textit{et al.} \cite{li2025abstract} instead constrain generation with predefined structures and apply post-processing verification and correction.
For verification, Huang \textit{et al.} \cite{huang2024llms} use solver outputs to evaluate generated models and introduce MAMO, which extends NL4OPT with ordinary differential equations and more complex LP and MILP problems.
Wang \textit{et al.} \cite{wang2024optibench} represent models as bipartite graphs and use a modified Weisfeiler-Lehman graph-isomorphism procedure to test equivalence in OptiBench.
Zhai \textit{et al.} \cite{zhai2025equivamap} further introduce Quasi-Karp equivalence and the EquivaMap system, which combine LLM-generated variable mappings with lightweight verification for scalable equivalence detection.

Prompt-based methods remain bounded by the capabilities and deployment constraints of their underlying LLMs.
Closed-source models such as the GPT series face two recurring limitations \cite{huang2025orlm}.
First, the scarcity of high-quality optimization-modeling data limits formal and domain-specific reasoning.
Second, API-based deployment can expose sensitive problem descriptions in applications such as medical scheduling and defense planning.
These limitations motivate learning-based methods that fine-tune open-source models to improve mathematical-formulation accuracy, preserve data control, and incorporate domain knowledge more directly.

\subsection{Learning-based Methods}
\label{Learning-based Methods}

Learning-based methods fine-tune model parameters so that optimization-modeling knowledge is encoded in the LLM rather than supplied only through prompts.
Early studies used fine-tuning for individual stages of the workflow.
Ner4OPT \cite{dakle2023ner4opt} combines conventional NLP methods with LLMs for NER, while Li \textit{et al.} \cite{li2023synthesizing} fine-tune a model to classify constraints and identify complex entity relations.
These models remain auxiliary components within two-stage pipelines.
LM4OPT \cite{ahmed2024lm4opt} moves toward direct model generation through progressive fine-tuning of Llama-2-7B \cite{touvron2023llama}.
However, because it is trained only on NL4OPT, its performance remains below that of closed-source models such as GPT-4 \cite{openai2023gpt4}.

ORLM \cite{huang2025orlm} established a general workflow that combines data synthesis, instruction tuning, and evaluation.
Its two-stage synthesis procedure produces aligned natural-language descriptions, mathematical models, and executable solver code.
The expansion stage uses seed examples and GPT-4 to generate problems across scenarios and difficulty levels.
The augmentation stage then modifies objectives and constraints and paraphrases problem descriptions, while matching correction and deduplication improve data consistency.
The resulting data are used to fine-tune open-source models, including Mistral-7B \cite{DBLP:journals/corr/abs-2310-06825} and DeepSeek-Math-7B-Base \cite{shao2024deepseekmath}.
These models outperform single-step GPT-4 generation and prompt-based methods such as CoE \cite{xiao2023chain} and OptiMUS \cite{ahmaditeshnizi2024optimus} in the reported evaluations.
Subsequent studies have largely followed this workflow while improving either data synthesis or fine-tuning.

One line of learning-based research focuses on synthesizing reliable and diverse training data \cite{lima2025toward}.
ReSocratic \cite{yang2024optibench} uses inverse generation: it first constructs mathematical optimization examples and then translates them into natural-language problem statements.
OptMATH \cite{lu2025optmath} uses bidirectional synthesis to control problem complexity and verifies generated pairs through forward modeling and rejection sampling.
DualReflect \cite{zhou2025auto} combines the diversity of forward generation with the reliability of inverse generation for dynamic programming problems.
Other work improves annotation quality and curriculum structure.
StructuredOR \cite{wang2024bpp} adds detailed modeling-process annotations, including variable definitions, while Step-Opt \cite{wu2025step} increases problem complexity iteratively and validates each generation step.

A parallel line of research improves the fine-tuning objective and training procedure \cite{jiang2024llmopt, chen2025solver, amarasinghe2023ai, zhou2025steporlm}.
LLMOPT \cite{jiang2024llmopt} uses multi-instruction fine-tuning for problem formalization and solver-code generation, together with alignment and self-correction mechanisms that reduce hallucinations.
SIRL \cite{chen2025solver} incorporates reinforcement learning and uses an external solver to provide verifiable rewards, which improves the factual and mathematical correctness of generated formulations.
Amarasinghe \textit{et al.} \cite{amarasinghe2023ai} adapt fine-tuning to business applications such as production scheduling and show that smaller, task-specific models can provide a cost-effective deployment option.

\subsection{Challenges}
\label{Challenges}

Current methods demonstrate that LLMs can map natural-language descriptions to formal optimization models, but they do not eliminate the central trade-offs of automated modeling.
CoE \cite{xiao2023chain} and ORLM \cite{huang2025orlm}, for example, can produce complete formulations and identify constraints that may be implicit in the original description.
However, prompt-based and learning-based methods differ substantially in deployment speed, data requirements, reliability, privacy, and maintenance cost.
Their principal limitations are summarized as follows:

\begin{itemize}
\item \textbf{Prompt-based methods} can be deployed rapidly without constructing a large annotated dataset.
However, their formulation accuracy is bounded by the formal reasoning and numerical capabilities of the pretrained model.
Errors commonly arise when the problem requires precise symbolic structure, implicit constraint recovery, or numerical consistency.
Reliance on external APIs also creates data-governance concerns in sensitive applications.
Private deployment can reduce exposure risk, but it does not resolve the underlying reasoning limitations.
Progress therefore requires stronger mathematical pretraining and hybrid architectures that connect LLM generation with symbolic parsers, solvers, or other external validators.

\item \textbf{Learning-based methods} improve reliability and data control by internalizing domain-specific modeling knowledge through synthetic data and fine-tuning.
However, data generation, verification, and retraining are expensive, particularly when they must be repeated across problem domains or model scales.
The field also lacks established procedures for updating fine-tuned models without performance regression.
Moreover, rapid improvements in general-purpose reasoning models can reduce the value of carefully engineered training pipelines.
Progress therefore requires data-efficient synthesis, verifiable training objectives, and sustainable model-maintenance protocols.
\end{itemize}

These trade-offs imply a deployment-dependent choice.
Prompt-based methods are appropriate when rapid deployment and minimal training data are the primary constraints.
Learning-based methods are preferable when reliability, privacy, and repeated use within a stable domain justify the training cost.
Section~\ref{Baseline Comparison and Guidance} evaluates this distinction empirically and converts it into practitioner-oriented selection guidance.

\section{LLMs for Optimization Solving}
\label{LLMs for Optimization Solving}

\begin{figure}[t]
\begin{center}
\includegraphics[width=0.45\textwidth]{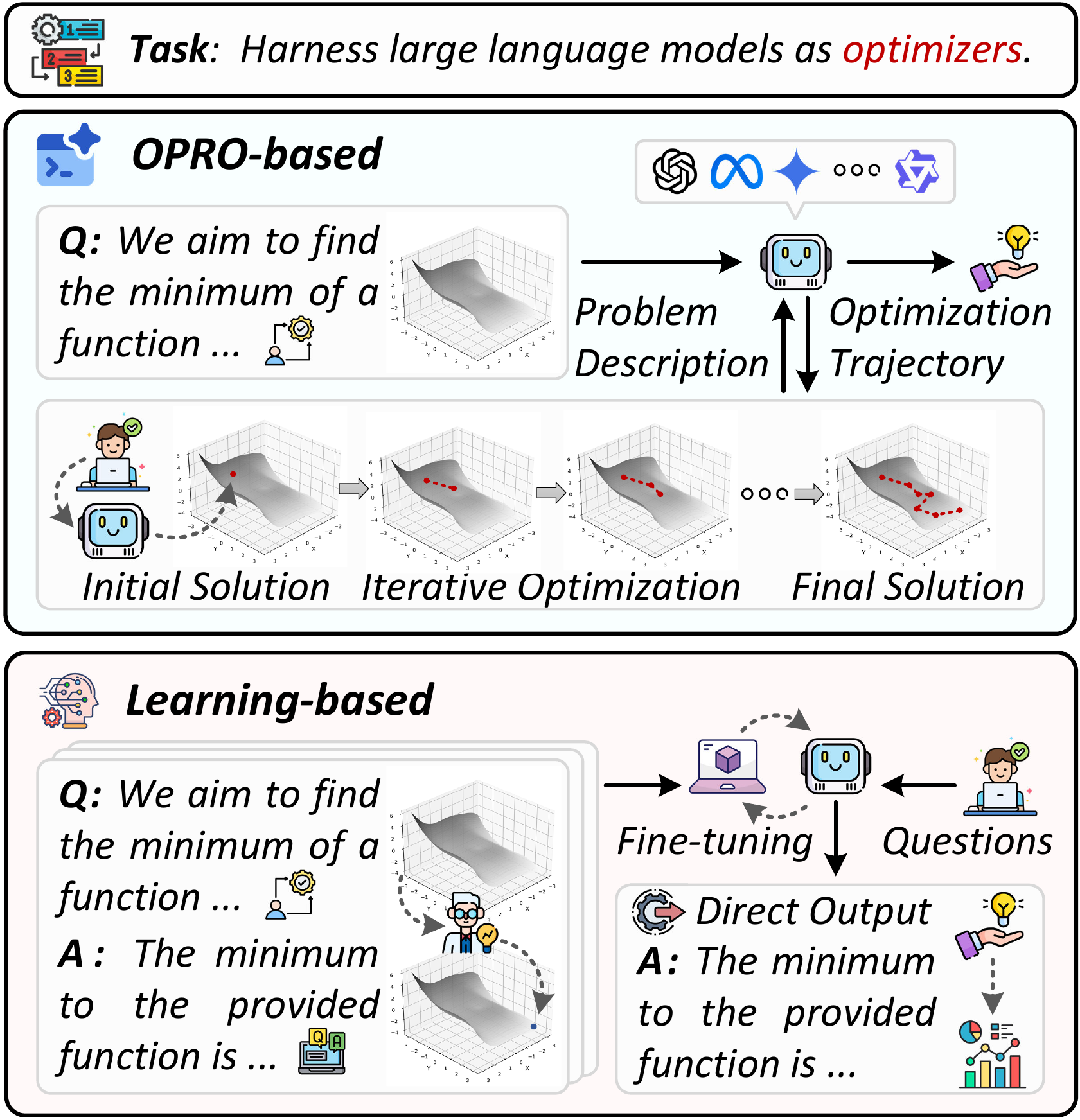}
\end{center}
\caption{Illustration of LLMs as optimizers. Most methods use iterative prompting, while a smaller group applies fine-tuning or pretraining to improve optimization performance.}
\label{fig:llmoptimizer}
\end{figure}

\begin{figure*}[ht!]
    \centering
    \includegraphics[width=0.90\textwidth]{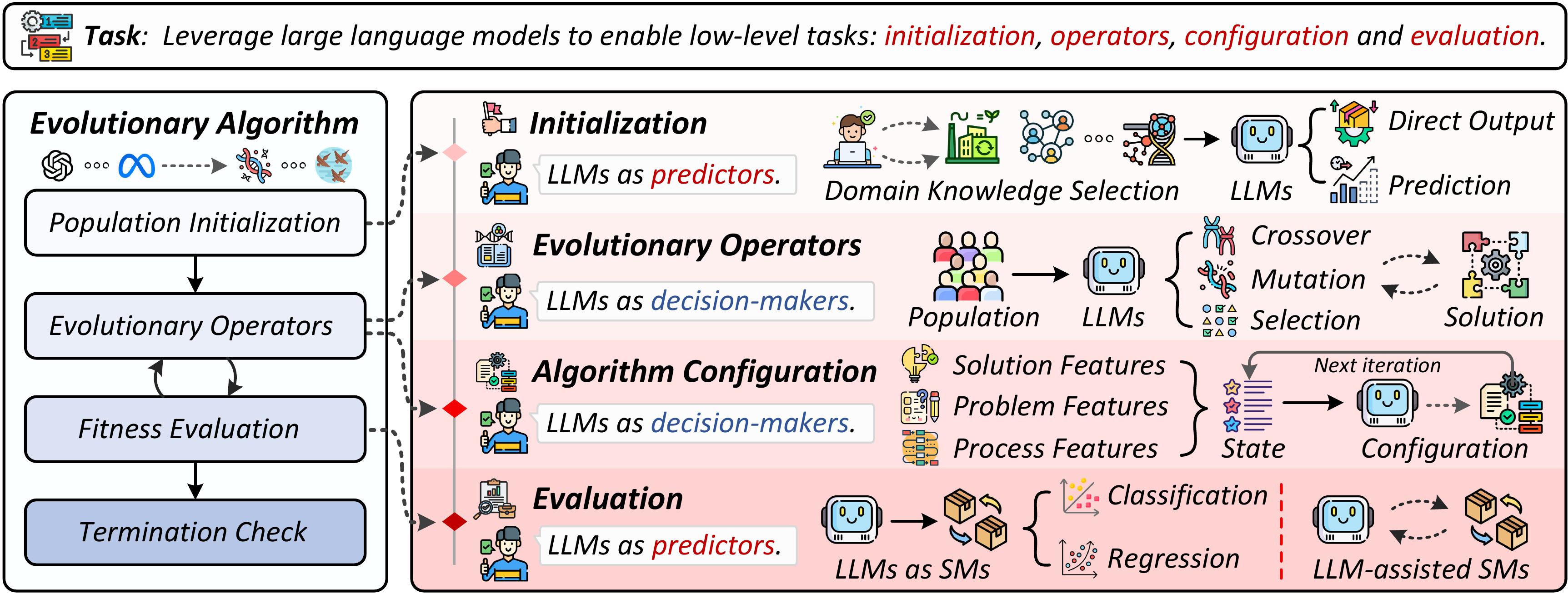}
    \caption{Illustration of low-level LLMs for optimization algorithms. LLMs can be applied at various stages within EAs, including initialization, evolutionary operators, algorithm configuration, and fitness evaluation.}
    \label{fig:low-level}
\end{figure*}

\begin{figure*}[ht!]
    \centering
    \includegraphics[width=0.90\textwidth]{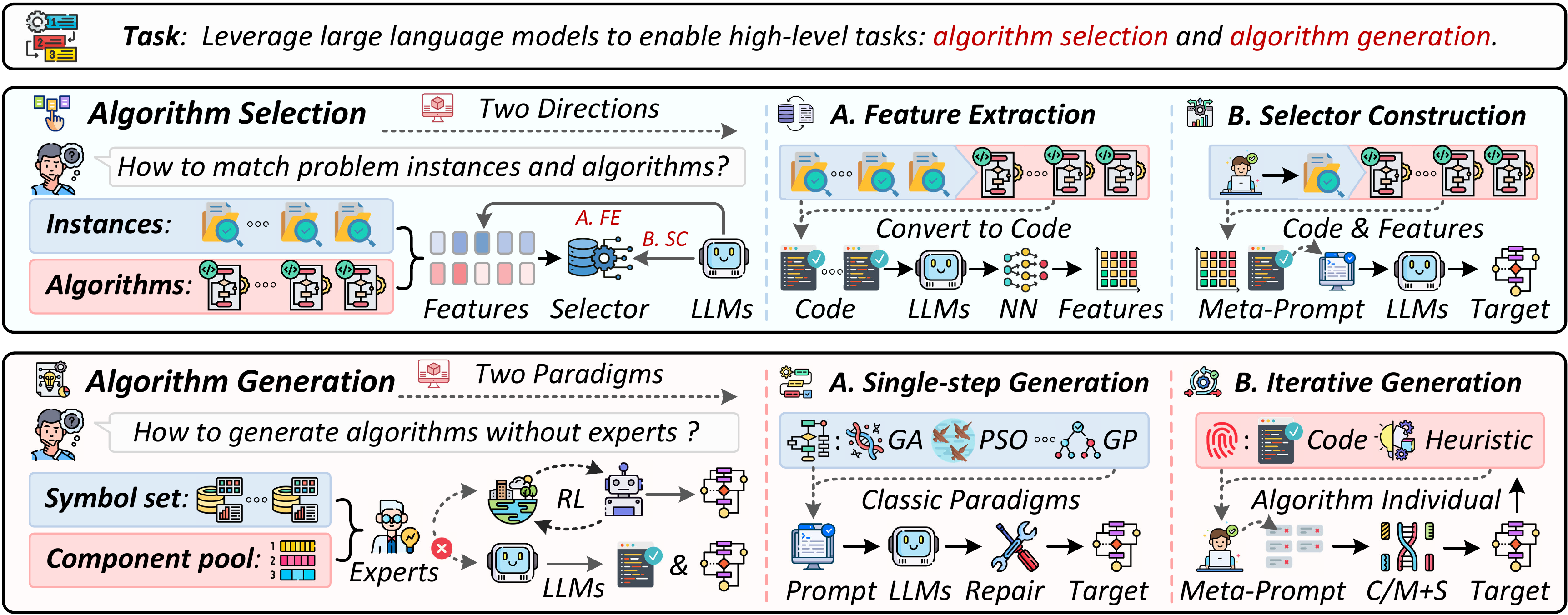}
    \caption{Illustration of high-level LLM-assisted optimization algorithms. Algorithm selection involves two key stages: feature extraction and selector construction. Algorithm generation has evolved from single-step to iterative generation, which reflects the shift toward more adaptive and context-aware design.}
    \label{fig:high-level}
\end{figure*}

Optimization solving is the execution stage in which an algorithm searches for solutions to a formulated problem.
Current research assigns LLMs three distinct roles in this stage: stand-alone optimizers, low-level components within optimization algorithms, and high-level systems for algorithm selection or generation.
Section~\ref{LLMs as Optimizers} reviews methods that use LLMs to generate solutions directly, as illustrated in Fig.~\ref{fig:llmoptimizer}.
Sections~\ref{Low-level LLMs for optimization Algorithms} and~\ref{High-level LLMs for optimization Algorithms} then examine component-level assistance and algorithm-level orchestration, as illustrated in Figs.~\ref{fig:low-level} and~\ref{fig:high-level}.
Section~\ref{Challenges2} compares the limitations of the three paradigms, and Supplementary Document II summarizes representative studies.

\subsection{LLMs as Optimizers}

\label{LLMs as Optimizers}

Stand-alone LLM optimizers use in-context learning to improve candidate solutions through iterative interaction.
OPRO \cite{yang2023large} established this paradigm by expressing an optimization problem in natural language and repeatedly prompting the LLM with previous solutions and their feedback.
The model infers patterns from these solution-feedback trajectories and proposes the next candidate.
This mechanism is most natural for discrete or language-structured tasks, where candidates can be represented semantically and inspected by a human.
Guo \textit{et al.} \cite{guo2023towards} evaluated the paradigm on gradient descent, hill climbing, grid search, and black-box optimization.
Subsequent applications include structural-matrix ordering \cite{jiang2024large}, wireless-network design \cite{qiu2024large}, hardware-workflow parameter tuning \cite{ghose2025orfs}, and jailbreak-attack analysis \cite{jiang2024optimizable}.

Later studies improved the information supplied during iteration rather than relying only on raw solution-feedback histories.
Lange \textit{et al.} \cite{lange2024large} provide sorted and discretized candidates as context, which reframes candidate generation as a sequence-recognition problem.
OPTO \cite{cheng2024trace} supplies structured execution traces, including intermediate values and function calls, to support a backpropagation-like update process.
Multimodal methods add visual structure to the prompt.
Huang \textit{et al.} \cite{huang2025multimodal} combine textual vehicle-routing descriptions with maps, Elhenawy \textit{et al.} \cite{elhenawy2024visual} combine scatter plots with multi-agent reasoning for the traveling-salesman problem, and Zhao \textit{et al.} \cite{zhao2025bridging} render graphs as images so that multimodal LLMs can access topological information.
These methods indicate that structured traces and visual representations can provide more useful optimization context than unstructured numerical histories.

Empirical studies nevertheless identify clear performance boundaries for stand-alone LLM optimization.
Zhang \textit{et al.} \cite{zhang2024revisiting} show that OPRO is strongly model-dependent and that smaller models have difficulty using long iteration histories.
Huang \textit{et al.} \cite{huang2024exploring} further compare LLMs on discrete and continuous black-box tasks.
Their results confirm relative strengths on symbolic and language-structured problems but expose weak performance on numerical continuous optimization.
The limitation becomes more pronounced when a problem lacks semantic structure, requires precise numerical calibration, has high dimensionality, or imposes a small evaluation budget.
Thus, the effectiveness of this paradigm depends strongly on how naturally the optimization state can be represented in the model's input space.

Two lines of work respond to these limitations.
The first attempts to improve stand-alone optimization through training.
Abgaryan \textit{et al.} \cite{abgaryan2024llms} fine-tune an LLM on instruction-solution pairs for single-step prediction, POM \cite{li2024pretrained} pretrains an optimizer for zero-shot transfer across black-box problems, and LLOME \cite{chen2024generalists} combines a two-level architecture with preference learning to generate valid sequences under biophysical constraints.
The second line changes the role of the LLM.
Rather than replacing an optimization algorithm, it embeds the LLM within an established framework and assigns it a narrower decision task.
This alternative motivates the low-level assistance paradigm reviewed next.

\subsection{Low-level LLMs for Optimization Algorithms}
\label{Low-level LLMs for optimization Algorithms}

Low-level assistance exploits the complementary strengths of population-based search and language-model reasoning \cite{chao2024large}.
EAs provide explicit mechanisms for exploration, selection, and constraint handling, while LLMs contribute semantic priors, domain knowledge, and generative decisions.
Embedding an LLM within an EA therefore preserves the algorithmic search structure and restricts the LLM to tasks for which its knowledge or reasoning is useful.
We organize this literature around four stages of an optimization algorithm: initialization, evolutionary operators, algorithm configuration, and evaluation.

\subsubsection{\textbf{Initialization}}
\label{Initialization}

Initialization determines the starting distribution of the search and can strongly affect convergence and final solution quality.
LLMs can inject prior knowledge by proposing candidates that are plausible before any objective evaluations are available.
In neural architecture search (NAS) \cite{xue2021self, huang2025exploring}, Yu \textit{et al.} \cite{yu2023gpt} use LLMs to propose architectural components, while Jawahar \textit{et al.} \cite{jawahar2023llm} use an LLM as a predictor to guide initial candidate selection.
In bioengineering, Teukam \textit{et al.} \cite{teukam2024integrating} generate mutant libraries for enzyme design, and De \textit{et al.} \cite{de2023optimized} generate initial portfolios for financial optimization.
However, initialization through an LLM incurs inference cost and does not guarantee feasibility.
Zhao \textit{et al.} \cite{zhao2025can} show that these limitations become more severe under strict constraints and at larger problem scales.

\subsubsection{\textbf{Evolutionary Operators}}
\label{Evolutionary Operators}

Evolutionary operators determine how an EA transforms existing solutions into new candidates.
Before LLMs, neural models such as attention and feed-forward networks were trained to approximate individual operators or complete algorithms \cite{lange2024evolution}.
Iterative-prompting methods such as OPRO \cite{yang2023large} suggested a training-free alternative: an LLM can generate variations after reading the problem description and selected parent solutions.
This formulation allows the operator to use semantic and domain information that is difficult to encode in conventional variation rules.

Several frameworks implement this idea at different levels of the evolutionary cycle \cite{huang2024advancing}.
LMX \cite{meyerson2024language} uses few-shot prompting for crossover and recombination of text-based genomes and generates semantically coherent offspring without additional training.
LMEA \cite{liu2024large} prompts LLMs to perform selection, crossover, and mutation and adapts the sampling temperature to balance exploration and exploitation.
Other methods specialize the role further: PAIR \cite{ali2025pair} focuses on selection, LMPSO \cite{shinohara2025large} adapts LLM guidance to particle swarm optimization, and LEO \cite{brahmachary2025large} uses LLM-generated strategies to regulate exploration and exploitation.

Operator-level assistance has also been extended to multi-objective optimization.
Liu \textit{et al.} \cite{liu2025large} integrate zero-shot LLM prompting with MOEA/D \cite{zhang2007moea} and use the observed behavior to inform the design of efficient white-box operators.
Because repeated LLM calls are expensive, hybrid methods invoke the model selectively.
Wang \textit{et al.} \cite{wang2024largeicic} use LLMs to generate only 10\% of each population, while conventional operators generate the remainder.
Liu \textit{et al.} \cite{liu2024largemulti} invoke an LLM only after population improvement becomes insufficient and otherwise rely on NSGA-II \cite{deb2002fast}.
These studies show that selective invocation can retain useful semantic variation while controlling computational cost.

\subsubsection{\textbf{Algorithm Configuration}}
\label{Algorithm Configuration}

Algorithm configuration determines hyperparameters and operator choices that strongly influence EA performance.
Conventional research distinguishes offline tuning from online control.
Recent MetaBBO methods formulate online control as a Markov decision process \cite{puterman1990markov} and learn policies through reinforcement learning \cite{chen2020self, pei2024learning, zhang2024improved}.
However, these policies usually require problem-specific training and may generalize poorly outside the training distribution.
LLMs offer a training-free alternative that maps observed search features to configuration decisions, although their ability to control numerical parameters remains uncertain.

Existing studies apply LLMs to both static tuning and dynamic control.
Kramer \textit{et al.} introduce LLM-based feedback loops for static parameter tuning \cite{kramer2024large} and dynamic control \cite{kramer2024llama} in evolution strategies.
Custode \textit{et al.} \cite{custode2024investigation} feed optimization trajectories to an OPRO-style controller for step-size adaptation and ask the model to justify each update.
LAOS \cite{zhang2025laos} reduces the redundancy of full trajectories by constructing a meta-prompt from features of the solution space, decision space, and search process; it then combines these features with prior optimization knowledge for adaptive operator selection.
Algorithm-generation frameworks such as EoH \cite{liu2024evolution} and LLaMEA \cite{van2024llamea} also tune parameters, but their primary objective is to generate algorithms rather than configure a fixed EA.

\subsubsection{\textbf{Evaluation}}
\label{Evaluation}

Evaluation is a major computational bottleneck when objective functions require expensive simulations or experiments.
Surrogate-assisted optimization reduces this cost by predicting objective values or candidate quality from previously evaluated solutions.
LLMs have been studied both as direct surrogates and as managers of surrogate models.
Hao \textit{et al.} \cite{hao2024large} formulate model-assisted selection as classification and regression and use historical observations to evaluate new candidates without additional training.
LICO \cite{nguyen2024lico} uses LLM representations to address data scarcity in molecular optimization.
At the management level, Rios \textit{et al.} \cite{rios2024large} use LLMs to support surrogate selection and training for engineering problems, while LLM-SAEA \cite{xie2025large} coordinates multiple expert agents to select surrogate models and infill criteria dynamically.

LLM-based surrogates have further been extended to multitask optimization.
Zhang \textit{et al.} \cite{zhang2025large} represent task observations as token sequences and use an LLM as a meta-surrogate to transfer information across tasks.
Together, these studies show that LLMs can support evaluation either by predicting candidate quality or by coordinating conventional surrogate models.
However, their robustness, calibration, and scalability require further evaluation before they can replace established surrogate-learning methods.

\subsection{High-level LLMs for Optimization Algorithms}
\label{High-level LLMs for optimization Algorithms}

High-level assistance assigns the LLM decisions that affect the optimization algorithm as a whole rather than individual search operations.
The LLM either selects an existing algorithm for a problem instance or generates an algorithm tailored to the task.
This role allows the model to reason over problem descriptions, algorithm properties, and previous performance records at a broader level.
Accordingly, this section reviews two forms of high-level assistance: \textit{algorithm selection} and \textit{algorithm generation}.

\subsubsection{\textbf{Algorithm Selection}}
\label{Algorithm Selection}

Algorithm selection identifies the most suitable method from a portfolio for a given problem instance.
The task is necessary because algorithms exhibit different strengths across problem structures and application domains.
Classical selection systems formulate it as a machine-learning problem with two stages \cite{wu2025towards}.
The first stage extracts statistical or landscape features that characterize the problem and candidate algorithms.
The second stage maps these features to a decision through classification \cite{abdulrahman2018speeding}, regression \cite{xu2008satzilla}, or hybrid models \cite{fehring2022harris}.
The reliability of the final selection therefore depends on both feature quality and the selector model.

LLMs have been introduced into both stages of this workflow.
AS-LLM \cite{wu2023large} uses code understanding to extract high-dimensional representations from source code or textual descriptions.
A feature-selection module retains the most relevant dimensions, separate networks encode algorithms and problems, and a similarity module produces the final selection.
InstSpecHH \cite{zhang2025llm} instead emphasizes the selector.
It first filters candidate algorithms by Euclidean distance and then asks an LLM to compare natural-language descriptions of the problem and the remaining algorithms.
Other systems select mathematical solvers inside broader modeling workflows \cite{ahmaditeshnizi2024optimus, ahmaditeshnizi2024optimus03}, but this task differs from classical portfolio-based algorithm selection and is treated separately in this survey.

\subsubsection{\textbf{Algorithm Generation}}
\label{Algorithm Generation}

Algorithm generation constructs an optimization algorithm from problem characteristics rather than selecting from a fixed portfolio.
Earlier reinforcement-learning methods reduce manual design but still depend on predefined search spaces \cite{ma2025toward}.
GSF \cite{yi2022automated} searches within a generic EA template whose component pools are specified manually, while SYMBOL \cite{chen2024symbol} removes the component pools but retains a manually defined symbol set.
LLMs can generate algorithmic concepts and executable code without an explicit component library, which expands the design space but also makes validation more difficult.

Research has progressed from single-step generation to iterative search \cite{wu2023large}.
In early demonstrations, Pluhacek \textit{et al.} \cite{pluhacek2023leveraging} prompt GPT-4 to decompose and recombine six metaheuristics, while Zhong \textit{et al.} \cite{zhong2024leveraging} generate the ZSO metaheuristic with GPT-3.5 and prompt engineering.
These studies establish feasibility but rely almost entirely on the prior knowledge and one-shot reasoning of the underlying model.
This dependence motivates iterative frameworks that evaluate generated algorithms and feed performance information back into the generation process.

Iterative generation now forms the main line of algorithm-generation research.
FunSearch \cite{romera2024mathematical} combines LLM generation with evolutionary selection to improve program fragments in function space.
AEL \cite{liu2023algorithm} and EoH \cite{liu2024evolution} extend this idea by co-evolving heuristic concepts and executable implementations.
EoH represents a heuristic as a natural-language idea and asks the LLM to produce corresponding code.
This dual representation permits search in an abstract semantic space while retaining executable evaluation, which can be more efficient than varying source code alone.
Reported experiments show that EoH generates heuristics that outperform established handcrafted baselines on several combinatorial benchmarks.

This line of work has produced supporting tools and extensions.
LLM4AD \cite{liu2024llm4ad} provides a platform for LLM-assisted algorithm design, while subsequent studies characterize the rugged and multimodal fitness landscape of the search \cite{liu2025fitness} and fine-tune LLMs through diversity-aware ranking, sampling, and direct preference optimization \cite{liu2025fine}.
MEoH \cite{yao2025multi} extends EoH to multi-objective heuristic search, and EoH-S \cite{liu2025eoh} co-evolves complementary algorithm sets.
Applications include edge-server scheduling \cite{yatong2024ts}, Bayesian optimization \cite{yao2024evolve}, and adversarial-attack design \cite{guo2024autoda}.

The LLaMEA series \cite{van2024llamea, van2025llamea} follows a complementary code-centered trajectory.
It uses an LLM as a variation operator within a conventional evolutionary loop and refines source code according to performance measurements and runtime feedback.
Unlike EoH, it does not explicitly co-evolve natural-language concepts and code.
The authors analyze the resulting search through code-evolution graphs \cite{van2025code} and behavior-space representations \cite{van2025behaviour}, and they introduce BLADE \cite{van2025blade} for standardized evaluation.
LLaMEA-HPO \cite{van2024loop} adds hyperparameter optimization to reduce iteration cost.
The framework has also been applied to Bayesian optimization \cite{li2025llamea} and photonic-structure design \cite{yin2025optimizing}.

Other iterative frameworks explore alternative search and reflection mechanisms \cite{sun2024autosat, huang2025autonomous, zhang2024understanding, yin2025controlling, li2025llm, ling2025complex}.
ReEvo \cite{ye2024reevo} treats the LLM as a hyper-heuristic and uses reflective evolution to revise candidate strategies.
MCTS-AHD \cite{zheng2025monte} applies Monte Carlo Tree Search to the heuristic design space to reduce premature convergence.
HSEvo \cite{dat2025hsevo} combines harmony search with genetic algorithms and uses diversity measures to balance exploration and exploitation.
Collectively, these methods mark a transition from isolated one-shot demonstrations to search frameworks that generate, evaluate, and revise algorithms iteratively.

\subsection{Challenges}
\label{Challenges2}

The three solving paradigms exhibit different strengths because they assign the LLM decisions at different structural levels.
Stand-alone methods such as OPRO \cite{yang2023large} ask the model to generate solutions directly.
Low-level methods such as LMEA \cite{liu2024large} and LMX \cite{meyerson2024language} restrict the model to particular operations within an EA.
High-level methods such as EoH \cite{liu2024evolution} and LLaMEA \cite{van2024llamea} search over algorithm designs.
This structural distinction produces three corresponding bottlenecks:

\begin{itemize}

\item \textbf{LLMs as Optimizers}: The main limitation is a mismatch between autoregressive sequence prediction and numerical search.
Stand-alone methods must encode the optimization state as a sequence and infer useful updates from solution-feedback histories.
Long, unstructured histories disperse attention and can obscure important intermediate states \cite{zhang2024revisiting, huang2024exploring}.
The problem becomes more severe for population-based search because the context must represent many candidates and their relations.
\begin{majorrevision}
Our BBOB experiments show that performance degrades sharply as dimensionality increases and that LLM optimizers stagnate even on the Sphere function.
\end{majorrevision}
Thus, this paradigm is most suitable for discrete or language-structured tasks and remains unreliable for high-precision, high-dimensional continuous optimization.
Future work should compress trajectories into structured state summaries or reposition the LLM as a problem-understanding module within a numerical optimizer.

\item \textbf{Low-level LLMs for Optimization Algorithms}: The main limitation is the scale and interdependence of the assigned decision space.
LLMs can be effective for local, discrete operations, but global tasks such as population initialization and full algorithm configuration require many coordinated decisions across combinatorial or continuous variables.
Autoregressive generation provides no explicit guarantee of global feasibility or consistency.
LLM controllers also tend to make conservative parameter updates, which can reduce exploration \cite{custode2024investigation, zhao2025can}.
\begin{majorrevision}
Our parameter-control experiments further indicate that discrete action representations produce more effective exploration than direct continuous outputs.
\end{majorrevision}
Low-level assistance is therefore most reliable when the decision scope is narrow and feedback is immediate.
Future methods should use hierarchical decomposition to convert global decisions into constrained subproblems that an LLM can address separately.

\item \textbf{High-level LLMs for Optimization Algorithms}: Algorithm generation searches a semantic design space that combines heuristic concepts with executable code, typically through EC or Monte Carlo Tree Search.
This process faces two main limitations.
First, it is computationally expensive because each candidate may require an LLM call, code execution, and evaluation on multiple benchmark instances.
Second, hierarchical design remains difficult: the model must map a problem description to an algorithmic framework, operators, and parameter settings while preserving executability and constraints.
Current methods therefore tend to recombine established strategies more often than they produce fundamentally new algorithmic principles.
Future work should develop surrogate evaluators for lower-cost search and connect LLMs with formal reasoning and verification modules for hierarchical, constraint-aware synthesis.
\end{itemize}

\begin{majorrevision}
These limitations support a paradigm-aware deployment principle.
Classical EAs remain the default for numerical and high-dimensional optimization.
LLMs are more useful when the task contains substantial language or semantic structure, or when the model is assigned a narrow decision role within an established algorithm.
Section~\ref{Baseline Comparison and Guidance} evaluates this principle empirically and translates it into method-selection guidance.
\end{majorrevision}

\begin{majorrevision}
\section{Empirical Study and Guidance}
\label{Empirical Study and Guidance}

To ground the preceding taxonomy empirically, this section systematizes the benchmark landscape and compares representative methods across the modeling-to-solving workflow.
Section~\ref{Benchmarks and Evaluation Protocols} reviews benchmarks and evaluation protocols for both modeling and solving.
Section~\ref{Baseline Comparison and Guidance} then combines collected results with controlled experiments on optimization modeling, stand-alone LLM optimization, low-level assistance, and high-level algorithm generation.
Figs.~\ref{fig:modeling_avg}, \ref{fig:optimizer}, \ref{fig:low_level_result}, and \ref{fig:high_avg} summarize the corresponding evidence.

\subsection{Benchmarks and Evaluation Protocols}
\label{Benchmarks and Evaluation Protocols}

Benchmark development has accompanied the rapid growth of LLM-based optimization methods.
A useful benchmark must satisfy two requirements: it should contain sufficiently diverse problems to expose method limitations, and it should support centralized evaluation under comparable protocols.
We review the benchmark landscape separately for optimization modeling and optimization solving.

Optimization-modeling benchmarks have developed through three stages.
The first was competition-driven and centered on NL4Opt \cite{ramamonjison2023nl4opt}.
The second expanded coverage through manually curated datasets, including MAMO \cite{huang2024llms}, NLP4LP \cite{ahmaditeshnizi2024optimus03}, NL2OPT \cite{mostajabdaveh2024optimization}, and ComplexOR \cite{xiao2023chain}.
The third introduced scalable synthetic generation, as represented by IndustryOR \cite{huang2025orlm}, OptiBench \cite{yang2024optibench}, OptMath \cite{lu2025optmath}, DP-Bench \cite{zhou2025auto}, and DCP-Bench \cite{michailidis2025cp}.
NL4Opt, MAMO, NLP4LP, ComplexOR, IndustryOR, OptiBench, and OptMath are among the most widely used datasets in recent studies \cite{liu2026optitree, zhou2025steporlm, xiao2026deepor}.
Supplementary Document III provides further dataset details.

Evaluation protocols fall into objective-wise and model-wise families.
Objective-wise evaluation executes the generated formulation with a solver and compares the resulting objective value with the ground truth.
CoE \cite{xiao2023chain} introduced this test-driven protocol, and pass@1 has become the most common metric.
However, a correct objective value does not guarantee that the generated model is structurally correct.
Model-wise evaluation therefore compares formulations directly.
NL4Opt converts models into coefficient matrices, while later methods use graph-edit distance \cite{xing2024towards} or modified graph-isomorphism tests with theoretical guarantees \cite{wang2024optibench}.
These measures provide finer-grained correctness scores, but they require reference models that are costly to construct.
Consequently, pass@1 remains dominant because it is easy to compute and directly reflects downstream solver behavior.

\begin{figure}[t]
\begin{center}
\includegraphics[width=0.48\textwidth]{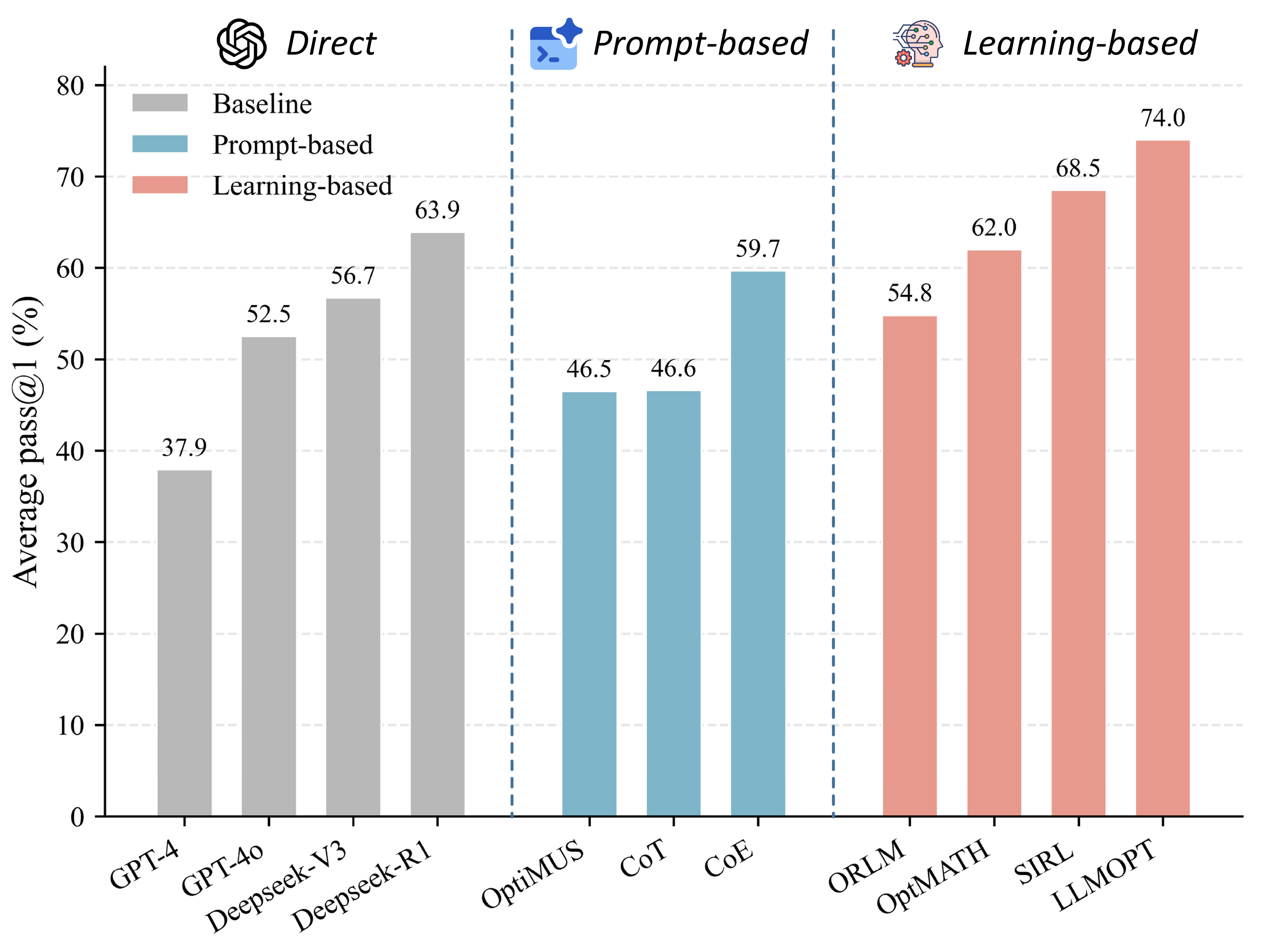}
\end{center}
\caption{Average pass@1 of representative methods from each category across all eight optimization modeling benchmarks. Methods are grouped as baseline, prompt-based, and learning-based approaches and are sorted within each group.}
\label{fig:modeling_avg}
\end{figure}

Benchmarks for optimization solving are less evenly distributed across the three paradigms.
Most existing suites focus on stand-alone LLM optimizers or high-level algorithm generation, while low-level assistance lacks a common benchmark.
For stand-alone optimization, NLGraph \cite{wang2023can}, PPLN \cite{aghzal2023can}, and GraphArena \cite{tang2025grapharena} primarily evaluate one-shot solutions on graph or language-structured tasks.
Opt-Bench \cite{li2025opt} extends this setting to iterative solving, in which the model revises candidates over multiple rounds.
Algorithm-generation benchmarks have developed more rapidly and include CO-Bench \cite{sun2026co}, FrontierCO \cite{feng2026frontierco}, HeuriGYM \cite{chen2025heurigym}, ALE-Bench \cite{imajuku2026ale}, and the LLM4AD platform \cite{liu2024llm4ad}.
Supplementary Document III summarizes these resources.

Evaluation of generated algorithms commonly combines three quantities.
The first is optimization quality relative to a best-known solution or an expert-tuned baseline.
The second is validity, which includes code executability and solution feasibility.
The third is cross-problem performance, which is often aggregated through normalized scores or rankings.
Although individual benchmarks implement these quantities differently, no single protocol has yet become standard.

\subsection{Baseline Comparison and Guidance}
\label{Baseline Comparison and Guidance}

We compare the four parts of the taxonomy using either curated results from common benchmarks or controlled experiments against established baselines.
For each part, we report the empirical pattern, identify its practical implication, and state the corresponding research opportunity.

\begin{figure}[t]
\begin{center}
\includegraphics[width=0.48\textwidth]{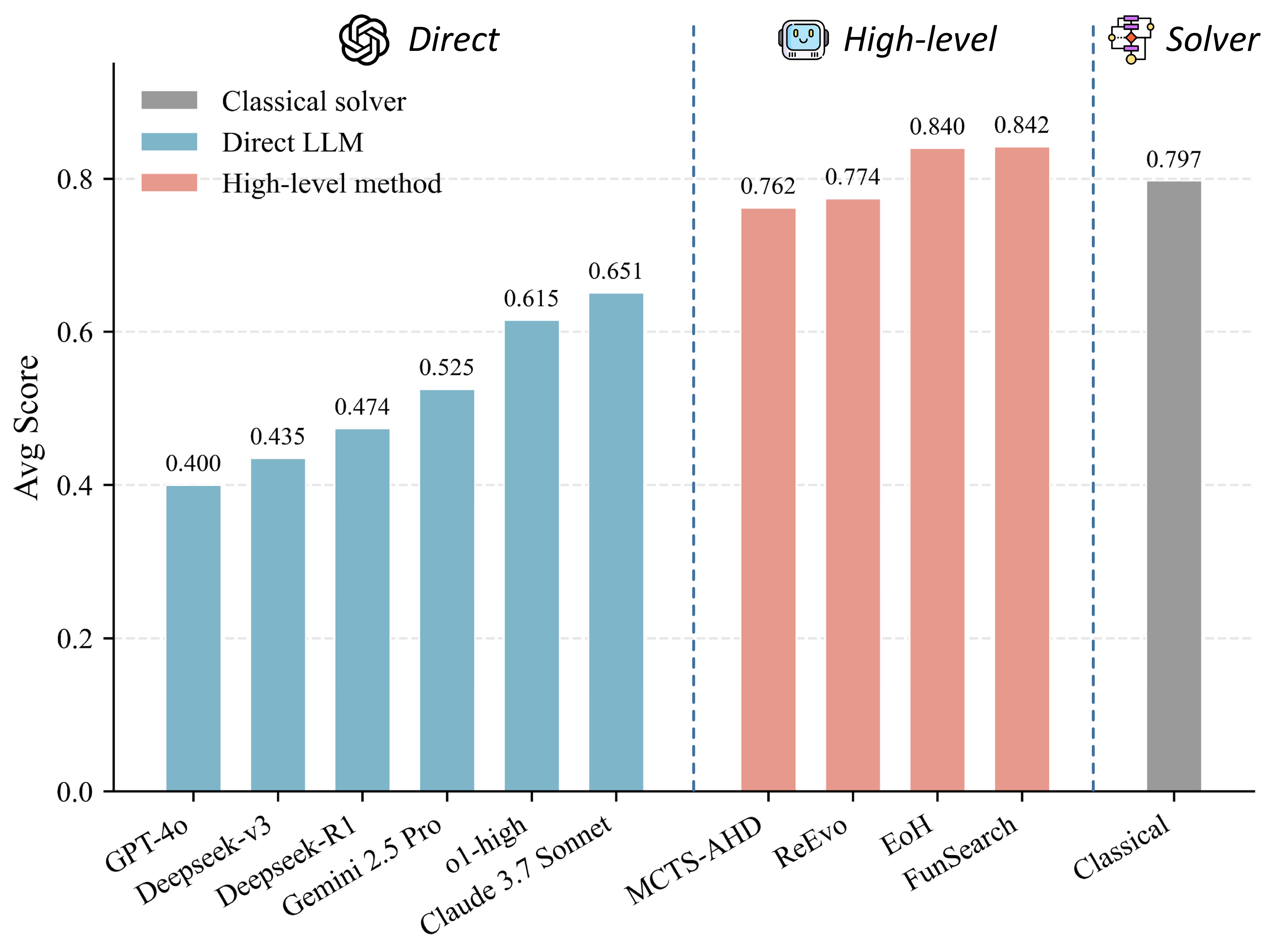}
\end{center}
\caption{Average normalized objective scores on CO-Bench for three groups: a well-tuned classical solver, direct LLM generation, and high-level agentic frameworks. Scores are averaged across all problems, with higher values indicating better performance and 1.0 matching the best-known solution.}
\label{fig:high_avg}
\end{figure}

\begin{figure}[t]
    \centering
    \begin{minipage}[t]{0.49\linewidth}
        \centering
        \includegraphics[width=\linewidth]{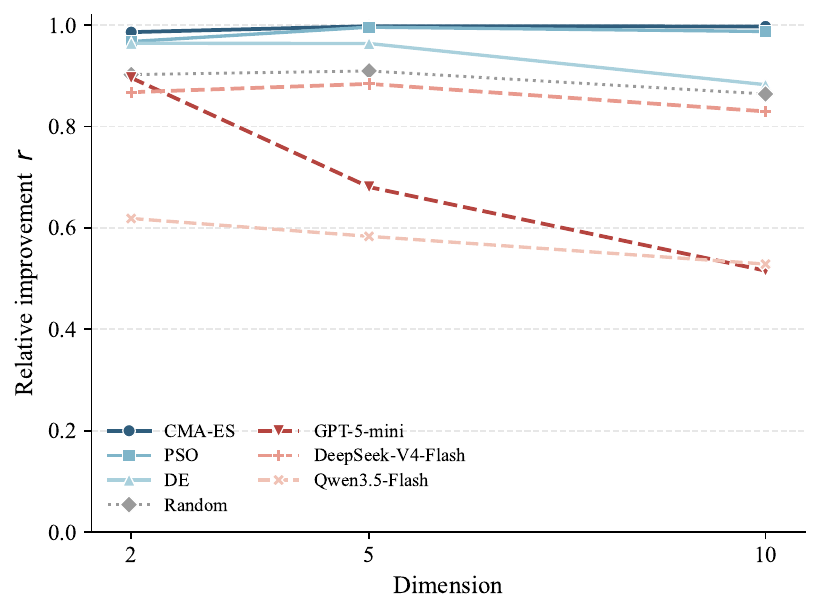}
        \vspace{1pt}
        \centerline{\small (a) Relative improvement}
    \end{minipage}
    \hfill
    \begin{minipage}[t]{0.49\linewidth}
        \centering
        \includegraphics[width=\linewidth]{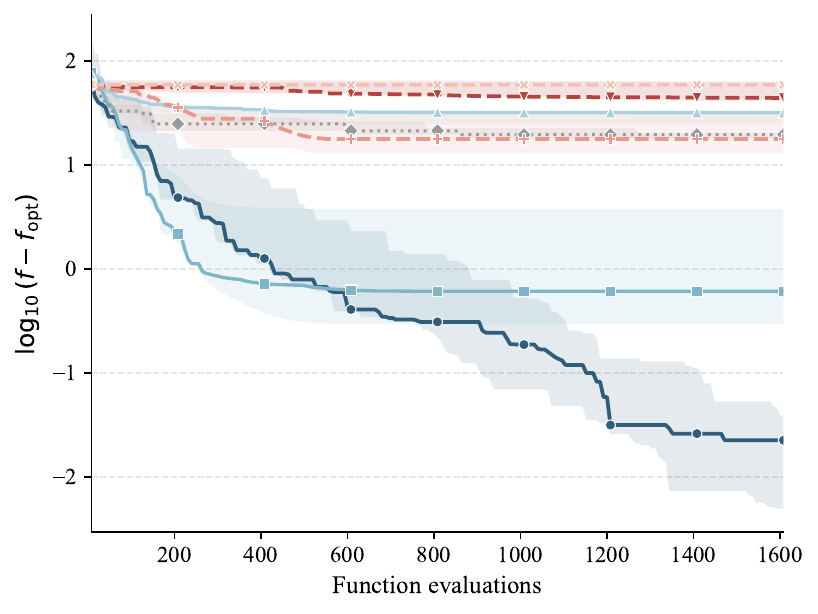}
        \vspace{1pt}
        \centerline{\small (b) Convergence on Sphere}
    \end{minipage}
    \caption{Performance comparison between LLM-as-optimizer and classical EAs on BBOB benchmarks. (a) Relative improvement $r$ averaged over four functions and five seeds; the results characterize how LLM-optimizer performance changes with dimensionality. (b) Median convergence trajectories on the 10-dimensional Sphere function; the trajectories reveal the early stagnation of LLM optimizers.}
    \label{fig:optimizer}
\end{figure}

\begin{figure}[t]
    \centering
    \begin{minipage}[t]{0.49\linewidth}
        \centering
        \includegraphics[width=\linewidth]{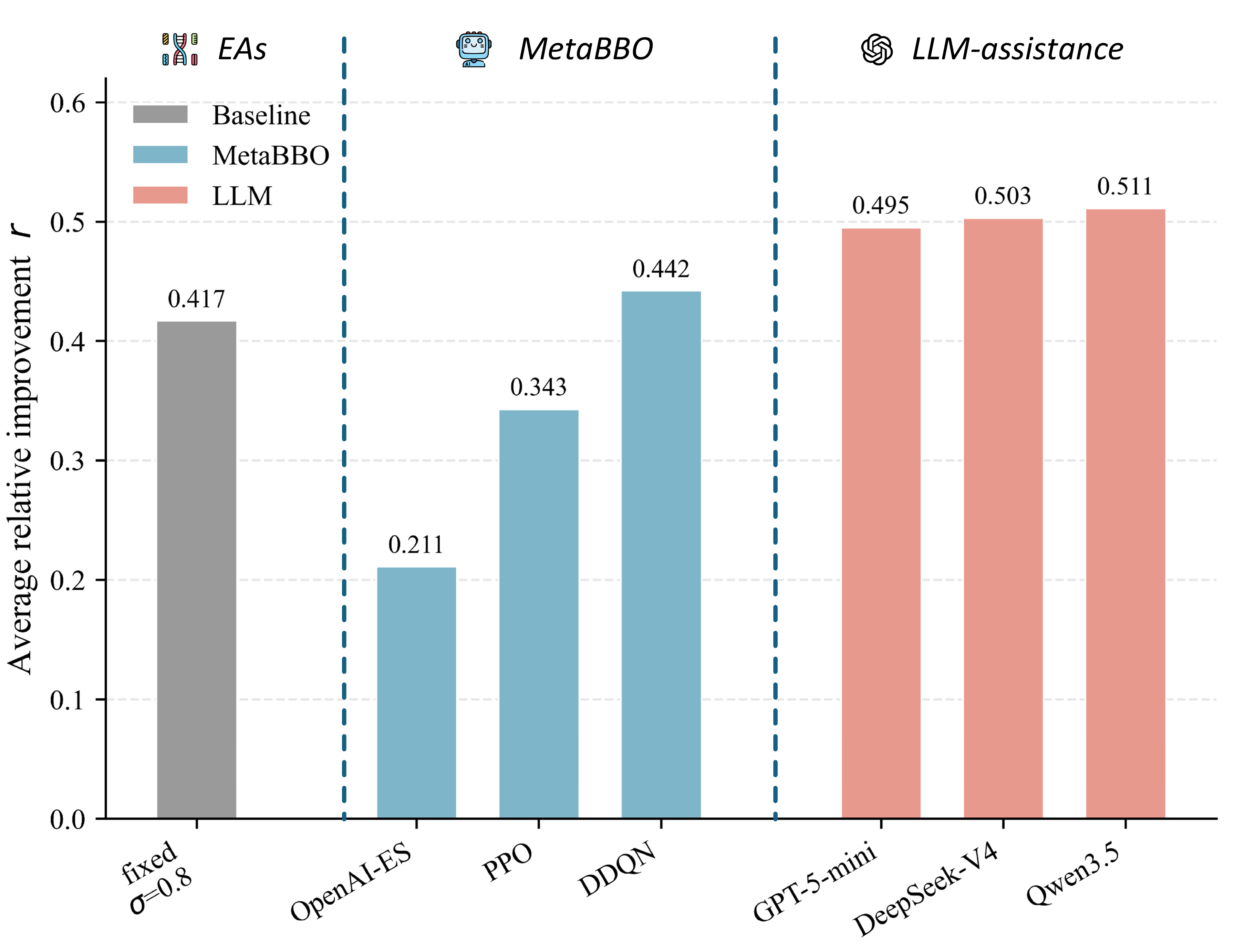}
        \vspace{1pt}
        \centerline{\small (a) Relative improvement}
    \end{minipage}
    \hfill
    \begin{minipage}[t]{0.49\linewidth}
        \centering
        \includegraphics[width=\linewidth]{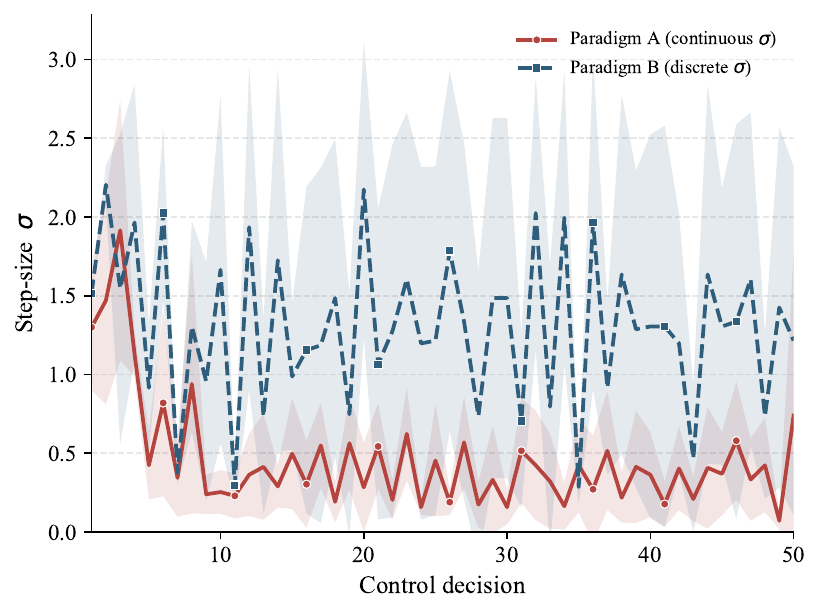}
        \vspace{1pt}
        \centerline{\small (b) Step-size trajectory}
    \end{minipage}
    \caption{Evaluation of LLM-driven step-size control in (1+1)-ES. (a) Average relative improvement $r$ across baseline, MetaBBO, and LLM controllers. (b) Step-size trajectory of Qwen3.5 on the Rastrigin function; the trajectory contrasts the exploitative behavior of continuous output with the exploratory behavior of discrete output.}
    \label{fig:low_level_result}
\end{figure}

For optimization modeling, we compare methods that report results on all eight selected benchmarks.
The baseline group contains GPT-4, GPT-4o, DeepSeek-V3, and DeepSeek-R1.
The prompt-based group contains OptiMUS \cite{ahmaditeshnizi2024optimus03}, chain-of-thought prompting \cite{wei2022chain}, and CoE \cite{xiao2023chain}, all implemented with GPT-4o.
The learning-based group contains ORLM \cite{huang2025orlm}, OptMATH \cite{lu2025optmath}, SIRL \cite{chen2025solver}, and LLMOPT \cite{jiang2024llmopt}.
Fig.~\ref{fig:modeling_avg} reports mean pass@1 across the eight benchmarks.
To maintain comparability, we include only methods evaluated on every benchmark and collect results from the original papers, benchmark reports, and recent comparative studies such as \cite{liu2026optitree}.
Supplementary Document IV provides the data-collection criteria, per-dataset results, and additional baselines.

The results favor learning-based methods in the current comparison.
Despite using open-source backbones with 8--32 billion parameters, LLMOPT and SIRL achieve average pass@1 scores of 74.0 and 68.5, respectively, compared with 63.9 for DeepSeek-R1.
Prompt-based performance is more variable: CoE reaches 59.7, whereas chain-of-thought prompting and OptiMUS remain near 46.5.
For practitioners who need a model without additional training, a strong closed-source LLM or an available specialized model such as LLMOPT is the most direct option.
When direct prompting is insufficient, search-augmented workflows such as CoE can improve accuracy without model training, but they increase API cost.
For researchers, the central question is whether specialized fine-tuning can maintain its advantage as general-purpose reasoning models improve, particularly when the open-source backbone limits the final performance.

To evaluate LLMs as numerical optimizers, we compare OPRO-style iterative optimization \cite{yang2023large} with classical methods on four BBOB functions \cite{hansen2021coco} across multiple dimensions and random seeds.
The classical baselines are random search, differential evolution (DE) \cite{das2010differential}, particle swarm optimization (PSO) \cite{kennedy1995particle}, and CMA-ES \cite{hansen2003reducing}.
The LLM optimizers use GPT-5-mini, DeepSeek-V4, and Qwen3.5 as backbones, and relative improvement $r$ is the primary metric.
Supplementary Document V reports the complete setup and results.

Fig.~\ref{fig:optimizer} shows the mean relative improvement and the median convergence trajectory on the 10-dimensional Sphere function.
The classical methods retain strong performance as dimensionality increases; CMA-ES and PSO both achieve $r>0.98$ at dimension 10.
The LLM optimizers perform worse and degrade with dimensionality.
For example, GPT-5-mini decreases from $r=0.896$ at dimension 2 to $r=0.516$ at dimension 10.
The Sphere trajectories further show early stagnation.
These results indicate that current iterative LLM optimization does not match mature EAs on numerical continuous problems.
Practitioners should therefore retain classical EAs as the default for this setting.
A more promising research direction is to convert raw trajectories into structured state summaries that preserve optimization-relevant information without requiring the model to reason directly over long vectors of floating-point values.

We use parameter control in a $(1+1)$-ES as a representative low-level task and compare LLM controllers with learning-based MetaBBO controllers.
DDQN \cite{sharma2019deep}, PPO \cite{ma2025toward}, OpenAI-ES \cite{lange2023discovering}, and the LLM controllers \cite{zhang2025laos} receive the same state observation.
The LLM group again uses GPT-5-mini, DeepSeek-V4, and Qwen3.5, with relative improvement $r$ as the evaluation metric.
We also compare continuous-output control \cite{custode2024investigation} with discrete-action control \cite{zhang2025laos} on Rastrigin to isolate the effect of the output representation.
Supplementary Document VI provides the full setup.

As shown in Fig.~\ref{fig:low_level_result}, all three LLM controllers outperform the fixed baseline and the trained MetaBBO controllers in this experiment.
This result contrasts with the weak performance of stand-alone LLM optimization and suggests that an LLM can be effective when assigned a small, well-defined decision problem.
The representation comparison further shows that discrete actions produce discernible exploration and exploitation, whereas direct continuous outputs remain concentrated within a narrow range.
For practitioners, LLM control is therefore a viable option for limited dynamic tuning, particularly when the action space can be discretized.
For researchers, the next step is to evaluate control at multiple granularities and to determine how population or variable decomposition can create decision units that remain compatible with LLM reasoning.

For high-level assistance, we focus on algorithm generation because it is the most active part of this paradigm.
Benchmark metrics remain heterogeneous, so we use CO-Bench \cite{sun2026co}, which compares direct LLM generation and agentic frameworks with a tuned classical solver under an average normalized objective score.
Supplementary Document VII relates these results to other studies and provides an error analysis.

Fig.~\ref{fig:high_avg} shows a clear distinction between one-shot generation and iterative agentic search.
The classical solver achieves an average score of 0.797, while the strongest direct-generation model, Claude 3.7 Sonnet, reaches 0.651.
Reasoning-oriented models consistently outperform their non-reasoning counterparts, but one-shot generation remains below the classical baseline.
In contrast, FunSearch \cite{romera2024mathematical} and EoH \cite{liu2024evolution} reach 0.842 and 0.840, respectively, while ReEvo \cite{ye2024reevo} and MCTS-AHD \cite{zheng2025monte} achieve 0.774 and 0.762.
The strongest agentic methods outperform the classical solver on more than half of the test instances.

This advantage is qualified by feasibility and novelty limitations.
Even the best agents have lower valid-solution rates than the classical solver, which reduces per-instance reliability.
Moreover, generated algorithms commonly recombine established techniques such as vectorization, local search, and simulated annealing rather than introduce new algorithmic principles.
For practitioners, agentic generation is competitive when iterative development and evaluation are affordable, whereas classical solvers remain preferable when feasibility guarantees are essential.
For researchers, the main opportunities are constraint-aware generation, lower-cost search, sustained improvement beyond fixed iteration budgets, and algorithmic novelty rather than more elaborate recombination of known components.

The empirical evidence supports a problem-driven selection rule rather than a universally best paradigm.
The first decision axis is problem structure.
Language-structured tasks, including mathematical formulation and combinatorial algorithm design, can benefit substantially from LLMs, whereas numerical and high-dimensional continuous problems remain better served by classical EAs.
The second axis is decision granularity.
LLMs are more effective for narrow and dynamic decisions, such as discrete parameter control, than for global or population-scale search.
The third axis is deployment constraints.
When feasibility, per-instance reliability, or data privacy is critical, classical solvers or locally deployed fine-tuned models are preferable to closed-source APIs.
Available training resources then determine whether prompt-based deployment or fine-tuning is appropriate.
Fig.~\ref{fig:guidance} consolidates these considerations into a method-selection flowchart.

\begin{figure*}[ht!]
    \centering
    \includegraphics[width=0.90\textwidth]{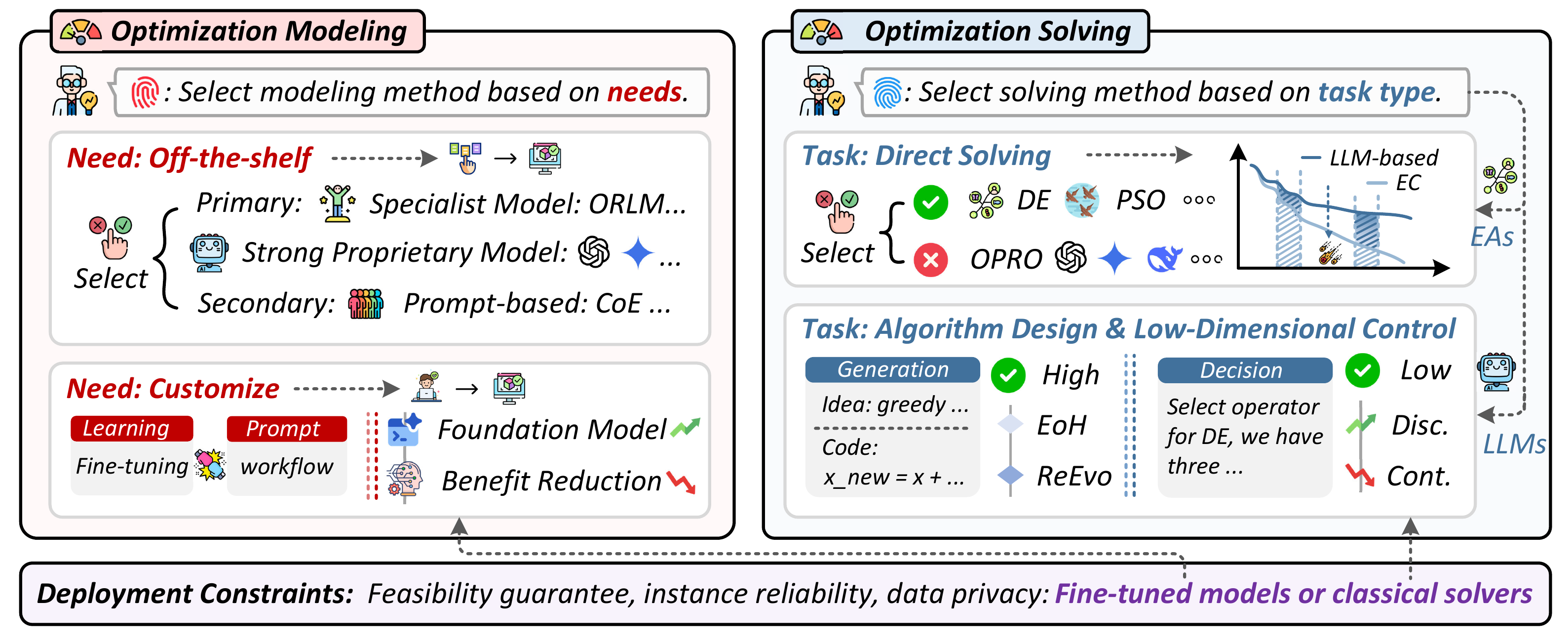}
    \caption{Method-selection flowchart that maps problem traits and deployment requirements to recommended LLM-for-optimization paradigms across the modeling route and the three solving paradigms.}
    \label{fig:guidance}
\end{figure*}

\end{majorrevision}
\section{Applications}
\label{Applications}

LLM-assisted optimization has been applied beyond benchmark numerical and combinatorial problems to tasks in computer science, the natural sciences, and engineering.
These applications differ in how they use the model: some rely on language-based problem formulation, some embed LLMs as operators or surrogates, and others use them to generate complete algorithms.
Section~\ref{Computer Science} reviews machine-learning and security applications.
Section~\ref{Natural Sciences} covers chemistry, biology, and physics.
Section~\ref{Engineering and Industry} examines engineering and industrial systems.

\subsection{Computer Science}
\label{Computer Science}

In computer science, neural architecture search (NAS) \cite{chen2023evoprompting} and security optimization \cite{guo2024autoda} are two representative application areas, while data augmentation, code optimization, and other tasks further demonstrate the breadth of the field \cite{wang2024llm, li2025cuda, ma2024llamoco, wang2025agenttts}.
NAS research mainly uses low-level assistance rather than stand-alone LLM optimization \cite{zhong2024large}.
Chen \textit{et al.} \cite{chen2023evoprompting} use LLMs as crossover and mutation operators within an EA \cite{yu2025llm}; few-shot prompts generate candidate architectures, and mixed sampling temperatures increase diversity.
This design has been extended to quality-diversity search with dual archives \cite{nasir2024llmatic}, role-based prompt diversification \cite{morris2024llm}, and graph neural architecture search \cite{mo2025autosgnn}.
LLMs also support initialization through pretraining \cite{yu2023gpt} or few-shot prediction \cite{jawahar2023llm}.

Security applications use LLMs both as direct optimizers and as search operators.
Jiang \textit{et al.} \cite{jiang2024optimizable} generate jailbreak suffixes through iterative self-reflection.
Other methods embed LLMs as heuristic operators for jailbreak-prompt optimization and improve efficiency or transferability \cite{li2025efficient, yu2024llm}.
AutoDA \cite{guo2024autoda} instead optimizes the generation function that produces adversarial examples, which allows attack strategies to evolve over repeated evaluations.

Related computer-science applications reformulate design artifacts as language-editable objects.
Wang \textit{et al.} \cite{wang2024llm} represent data-augmentation policies as natural-language instructions and apply crossover and mutation to adapt them to long-tailed data.
Li \textit{et al.} \cite{li2025cuda} iteratively generate and optimize CUDA kernels and use execution performance as feedback without manual code revision.

\subsection{Natural Sciences}
\label{Natural Sciences}

Natural-science problems in chemistry \cite{wang2025survey}, biology \cite{nana2025integrating}, and physics \cite{zhang2025autoturb} often combine high-dimensional search spaces, expensive evaluations, and domain knowledge that is difficult to encode explicitly.
LLMs are therefore used primarily to inject scientific priors, generate structured candidates, or coordinate expensive search rather than to replace numerical optimization entirely.

Molecular discovery is the main chemistry application of LLM-assisted optimization \cite{wang2025survey}.
Most methods embed chemically informed LLMs within EAs.
Wang \textit{et al.} \cite{wang2024efficient} use an LLM for molecular crossover and mutation.
Guevorguian \textit{et al.} \cite{guevorguian2024small} replace genetic operators with fine-tuned LLMs and use a predictive model to reduce stagnation; the approach is later extended to multi-objective molecular discovery \cite{ran2025mollm}.
At the evaluation level, LICO \cite{nguyen2024lico} uses LLM representations as surrogates when molecular data are scarce.

Biological applications also favor component-level integration.
LLM-GA \cite{nana2025integrating} uses an LLM to initialize mutant libraries for enzyme design.
Tran \textit{et al.} \cite{tran2024protein} and Wang \textit{et al.} \cite{wang2025large} use LLMs as mutation or crossover operators for protein design.
Chen \textit{et al.} \cite{chen2024llms, chen2024generalists} develop a two-level optimizer that iteratively modifies biological sequences under complex constraints.
ProLLaMA \cite{lv2025prollama} spans both modeling and solving by mapping natural-language descriptions to protein-function predictions and then optimizing the corresponding sequences.

Physics applications span modeling and solving across fluid dynamics \cite{zhang2025autoturb, zhang2025using}, semiconductor systems \cite{li2023english}, and statistical physics \cite{du2024large, sun2025integrating}.
At the modeling level, Du \textit{et al.} \cite{du2024large} use GPT-3.5 to derive partial differential equations from data, Zhang \textit{et al.} \cite{zhang2025autoturb} apply LLMs to turbulence-closure modeling, and Ma \textit{et al.} \cite{ma2024llm} connect modeling with solving in a unified workflow.
At the solving level, Li \textit{et al.} \cite{li2023english} use LLMs to generate and debug code for laser-parameter optimization, while Zhang \textit{et al.} \cite{zhang2025using} use LLMs to propose parameter combinations for fluid-dynamics simulations.

\subsection{Engineering and Industry}
\label{Engineering and Industry}

Engineering and industrial applications concentrate on wireless communications \cite{qiu2024large}, industrial design \cite{yin2025optimizing}, and edge-computing or scheduling problems \cite{yatong2024ts}.
These domains combine structured engineering knowledge with expensive simulation or deployment constraints, which creates opportunities for LLMs at both the modeling and solving stages.

Wireless-network research covers the complete workflow from formulation to search.
For modeling, LLM-OptiRA \cite{peng2025llm} identifies nonconvex components in resource-allocation problems and reformulates them into solvable forms.
Wen \textit{et al.} \cite{wen2025hybridrag} use retrieval-augmented generation to incorporate external expert knowledge into the formulation process.
For solving, Qiu \textit{et al.} \cite{qiu2024large} use LLMs directly for combinatorial tasks such as access-point placement, and related studies address resource allocation \cite{lee2024llm}, power control \cite{zhou2024large}, and multi-UAV deployment \cite{wang2025multi}.
Low-level methods embed LLMs in multi-objective optimization for integrated sensing and communication in UAV networks \cite{li2025large}.
The LHS framework \cite{hou2024wireless} combines heuristic recommendation, informed initialization, and iterative refinement.

Industrial-design applications use LLMs for configuration, direct search, and algorithm generation.
Ghose \textit{et al.} \cite{ghose2025orfs} develop an agent that tunes chip-design parameters for performance, power, and area.
Jiang \textit{et al.} \cite{jiang2024large} use LLMs to optimize design-structure matrices that encode dependencies within engineering systems.
Yin \textit{et al.} \cite{yin2025optimizing} apply LLaMEA \cite{van2024llamea} to generate algorithms for photonic-structure design.
Beyond design, Yatong \textit{et al.} \cite{yatong2024ts} apply LLM-generated heuristics to task scheduling in edge-server environments.

\section{Vision for the Field}
\label{Vision for the Field}

The reviewed literature reveals substantial methodological diversity, but it also exposes three structural gaps that limit further progress.
First, optimization modeling and solving remain weakly connected.
Second, most methods are designed offline and adapt only minimally during search.
Third, existing systems usually assign an LLM a single isolated role rather than organizing multiple agents, solvers, tools, and human experts into a persistent optimization ecosystem.
These gaps motivate three corresponding research directions: integrated modeling and solving, dynamic self-adaptation, and agentic coordination.

\subsection{Bridging Modeling and Solving}
\label{Bridging the Gap Between Modeling and Solving}

Optimization modeling and solving have largely developed as separate research lines.
OptiMUS \cite{ahmaditeshnizi2024optimus, ahmaditeshnizi2024optimus03} connects natural-language modeling with program generation through an agent-based conductor, while ORLM \cite{huang2025orlm} generates mathematical formulations and solver code through a fine-tuned model.
However, both systems ultimately rely on conventional external solvers.
The LLM therefore remains concentrated at the front end and cannot exploit advances in adaptive configuration, algorithm selection, or algorithm generation during the solving process.

Bridging the two stages requires both sequential integration and reciprocal interaction.
The first direction is an \emph{end-to-end LLM-driven workflow} that links problem understanding, formulation, algorithm design, execution, and evaluation within one closed loop.
\begin{majorrevision}
Such a system should not merely call a fixed external solver; it should select, configure, or generate solving strategies according to the formulated problem and observed search behavior.
The second direction is parallel co-evolution of the model and the algorithm.
Model-derived structures and constraint signatures can condition algorithm design, while search trajectories and failure modes can expose missing constraints or modeling errors and trigger reformulation.
This reciprocal feedback would turn modeling and solving into mutually corrective processes rather than a one-way pipeline.
\end{majorrevision}

\subsection{Transitioning from Static to Dynamic Methods}
\label{Transitioning from Static to Dynamic Methods}

Most current LLM-based optimization methods are \emph{static}: their prompts, algorithm structures, and interaction protocols are specified before the search and change little in response to online evidence.
Reinforcement-learning research provides a contrasting model through dynamic configuration \cite{tian2022deep, sharma2019deep}, dynamic selection \cite{guo2024deep}, and dynamic generation \cite{chen2024symbol}.
These methods adapt decisions to the current optimization state.
Dynamic selection schedules algorithms according to their observed strengths, while dynamic generation constructs operators for changing search conditions.

Dynamic selection and generation remain underexplored in LLM-based optimization.
Existing adaptive examples are mostly local, including interactive modeling \cite{lawless2024want} and parameter control \cite{custode2024investigation, zhang2025laos}.
A broader objective is the \emph{self-evolution of algorithms}, in which an LLM revises representations, operators, parameters, and coordination strategies according to search feedback.
\begin{majorrevision}
Such systems would retain evaluated experience, reuse successful components, and alter their own search logic as the problem or environment changes.
The central technical challenge is to support continual adaptation without uncontrolled drift, loss of feasibility, or repeated rediscovery of previously learned strategies.
\end{majorrevision}

\subsection{Toward an Agentic Ecosystem for Optimization}
\label{Toward an Agentic Optimization Ecosystem}

LLMs should ultimately be viewed not as isolated optimization engines but as agents within a larger optimization ecosystem.
In such a system, agents can collaborate with human experts, mathematical solvers, simulators, and domain tools.
They can capture tacit knowledge, such as driver-routing practices \cite{liu2023can} or scheduling constraints \cite{jobson2024investigating}, and support trustworthy interaction through infeasibility diagnosis \cite{chen2024diagnosing} and decision explanation \cite{singh2024enhancing, chacon2024large, maddigan2024explaining, kikuta2024routeexplainer}.

\begin{majorrevision}
Current multi-agent frameworks commonly rely on a centralized conductor \cite{xiao2023chain, li2024stride, lawless2024want, wang2025agenttts}, which can become a bottleneck as the number and diversity of agents increase.
EvoGit \cite{huang2025evogit} provides an alternative coordination model based on a shared phylogenetic graph and asynchronous, reward-free interaction.
Adapting such structure-driven coordination to optimization could allow agents to develop and compare algorithms through a shared partial-order representation without requiring a single controller to determine every interaction.

An agentic optimization ecosystem must also support adaptation and knowledge transfer.
Real-world objectives and constraints change over time \cite{lawless2024want}, so the system should detect distribution shift through performance degradation, constraint violations, or changes in user requirements.
It should then revise its modeling and solving strategies online. This process would extend interactive formulation toward continual task-aware optimization \cite{liu2026evox}.

Cross-domain transfer is equally important.
Although LLMs contain broad pretrained knowledge, most current optimization systems remain specialized to one domain or problem family.
Unified representations of optimization trajectories and modular skill libraries \cite{alzubi2026evoskill} could allow agents to retrieve, compose, and adapt experience across heterogeneous tasks.
The resulting ecosystem would treat algorithms, prompts, models, and domain procedures as reusable and evolvable components rather than isolated solutions.
\end{majorrevision}

\section{Conclusion}
\label{Conclusion}

This survey organized LLMs for optimization through a modeling-to-solving perspective.
For optimization modeling, we reviewed prompt-based and learning-based methods that translate natural-language descriptions into formal variables, objectives, constraints, and solver code.
For optimization solving, we distinguished three roles: stand-alone LLM optimizers, low-level LLM components embedded in established algorithms, and high-level LLM systems for algorithm selection and generation.
This taxonomy clarifies how the effectiveness of an LLM depends on the structural level and decision scope assigned to it.

\begin{majorrevision}
The benchmark analysis and baseline comparisons further identify consistent deployment boundaries.
Classical EAs remain the most reliable choice for numerical and high-dimensional continuous optimization.
LLMs provide greater value when the task contains substantial language or semantic structure, when they generate or select algorithms at a high level, or when they make narrow component-level decisions such as discrete parameter control.
Prompt-based modeling offers rapid deployment, while fine-tuned models can improve reliability when sufficient data and maintenance resources are available.

Future progress requires systems that connect modeling with solving, adapt their algorithms during search, and coordinate multiple agents, tools, solvers, and human experts.
Such systems should support reciprocal feedback between formulation and algorithm design, continual adaptation without loss of reliability, and transfer of reusable optimization knowledge across domains.
By identifying these requirements, this survey provides both a structured account of the current field and a basis for developing dynamic, self-evolving, and agentic optimization systems.
\end{majorrevision}

\bibliographystyle{IEEEtran}
\bibliography{llmea}

\clearpage
\appendices

\onecolumn

{\centering\Large\bfseries A Systematic Survey on Large Language Models for Evolutionary Optimization: From Modeling to Solving \\ (Supplementary Document)\par}
\vspace{1em}

\renewcommand{\appendixname}{}   

\section{Background}
\label{Background}

Research on LLM-enabled component control, high-level orchestration, and algorithm generation has focused primarily on evolutionary algorithms (EAs). Accordingly, this section provides the background required for the survey. Section~\ref{Evolutionary Algorithms} introduces the common framework and major paradigms of EAs, while Section~\ref{Large Language Models} summarizes LLM architectures and the principal techniques used to adapt them.

\subsection{Evolutionary Algorithms}
\label{Evolutionary Algorithms}

Evolutionary algorithms (EAs) are population-based, gradient-free optimization methods inspired by biological evolution. A typical EA begins by initializing a population of candidate solutions, whose representations may take the form of numerical vectors, permutations, program trees, or neural networks. At each iteration, a fitness function evaluates the candidates. Parent-selection and variation operators, such as crossover and mutation, then generate offspring, after which environmental selection constructs the next population from the parents and offspring. Many EAs also retain elite candidates to preserve the best solutions found so far. This cycle continues until a termination criterion is met, and the best-performing candidate is returned as the final solution.

Over several decades, this general framework has produced a broad range of established paradigms \cite{cheng2014competitive, tian2017platemo}, including Genetic Algorithms (GAs) \cite{holland1992genetic}, Genetic Programming (GP) \cite{koza1994genetic}, Differential Evolution (DE) \cite{das2010differential}, Particle Swarm Optimization (PSO) \cite{kennedy1995particle}, and Covariance Matrix Adaptation Evolution Strategy (CMA-ES) \cite{hansen2003reducing}. Although these paradigms share a population-based iterative structure, they employ distinct search mechanisms for different problem domains. GAs emulate natural selection through genetic operators and are well suited to discrete and combinatorial problems. DE and PSO are widely used for continuous optimization: DE generates offspring by adding weighted differences between population vectors, whereas PSO updates each particle according to its personal best and the population's global best. CMA-ES adapts the full covariance matrix of its mutation distribution and performs particularly well on ill-conditioned continuous landscapes. GP uses tree-structured representations to support automatic programming and symbolic regression.

Compared with exact optimization methods, EAs are often better suited to non-convex, multimodal, and noisy problems. Their applications range from aerospace design \cite{arias2012multiobjective} to hyperparameter optimization in deep learning \cite{elsken2019neural}. However, the No Free Lunch (NFL) theorem \cite{wolpert2002no} establishes that no optimizer is universally superior across all problem classes. Practitioners must therefore invest substantial effort in selecting, configuring, and designing appropriate EA variants. This challenge has motivated the integration of EAs with machine learning (ML) \cite{talbi2021machine}, including reinforcement-learning-based frameworks such as MOEA/D-DQN \cite{tian2022deep} and SYMBOL \cite{chen2024symbol}. More recently, researchers have investigated how LLMs can support EAs and optimization more broadly \cite{wu2025evolutionary}, from problem formulation \cite{huang2025orlm} to automated algorithm design \cite{liu2024systematic}.

\subsection{Large Language Models}
\label{Large Language Models}

Large language models (LLMs) are built primarily on the Transformer architecture \cite{vaswani2017attention}, which models global input-output dependencies through self-attention and supports parallel training on large corpora. Three principal architectural paradigms have emerged: encoder-decoder, encoder-only, and decoder-only models \cite{cai2021compare}. Encoder-decoder models, exemplified by BART \cite{lewis2019bart} and T5 \cite{raffel2020exploring}, follow the original Transformer design and are effective for sequence-to-sequence tasks. Encoder-only models, such as BERT \cite{devlin2019bert}, emphasize contextual representation learning for language understanding. Decoder-only models, including ChatGPT \cite{floridi2020gpt} and DeepSeek \cite{guo2024deepseek}, generate text autoregressively and dominate current research on LLMs for optimization because of their strong generation and reasoning capabilities. These models are commonly adapted through two routes: prompt engineering \cite{white2023prompt} and fine-tuning \cite{han2024parameter}.

Prompt engineering steers LLM outputs through carefully designed instructions without modifying model parameters. Zero-shot and few-shot prompting \cite{kojima2022large} use pretrained knowledge either directly or through a small number of in-context examples. Structured reasoning methods extend this approach. Chain-of-Thought (CoT) prompting \cite{wei2022chain} elicits intermediate reasoning steps for complex tasks, while Tree-of-Thought (ToT) \cite{yao2023tree} and Graph-of-Thought (GoT) \cite{besta2024graph} organize candidate reasoning paths as searchable trees or graphs. Retrieval-augmented generation (RAG) supplements the model's parametric knowledge with external information at inference time. Domain-specific prompting strategies, including Chain-of-Code (CoC) \cite{li2023chain} and Chain-of-Knowledge (CoK) \cite{li2023chain2}, further tailor the reasoning process to particular task structures.

Fine-tuning updates model parameters on task-specific data. This process enables deeper adaptation but requires additional labeled data and computational resources. Parameter-efficient fine-tuning (PEFT) methods, such as Low-Rank Adaptation (LoRA), reduce this cost by updating only a small subset of parameters while keeping the backbone fixed. Two widely used fine-tuning strategies are instruction tuning and preference alignment. Instruction fine-tuning \cite{liu2023visual} reformulates diverse tasks as instruction-input-output triplets and trains the model to follow those instructions. This approach improves zero-shot generalization, as demonstrated by Flan-T5 \cite{chung2024scaling}. Alignment fine-tuning seeks to make model behavior consistent with human intentions and preferences. Reinforcement Learning with Human Feedback (RLHF) \cite{ouyang2022training} optimizes a reward model derived from human judgments, whereas Direct Preference Optimization (DPO) \cite{rafailov2023direct} directly optimizes the likelihood of preferred outputs. Together, these techniques support the development of more reliable and useful LLMs.

\section{Representative Works\label{sec:rep}}

This section consolidates the detailed taxonomies of the studies reviewed in the main text. Table~\ref{tab:formulation} summarizes representative work on LLMs for optimization modeling, including each method's objective and methodological category. Table~\ref{tab:solving} catalogs studies on LLMs for optimization solving and classifies them according to the role of the LLM in the optimization workflow, such as direct optimizer, low-level assistant, or high-level algorithm generator.

\begin{table*}[h]
    \centering
    \caption{Representative studies on LLMs for optimization modeling, grouped into prompt-based and learning-based methods.}
    \label{tab:formulation}
    \renewcommand{\arraystretch}{1.2}
    \begin{adjustbox}{max width=\linewidth}
    \begin{tabular}{c|c|c|c|l}
    \toprule
    \textbf{} & \textbf{Method} & \textbf{Venue} & \textbf{Type} & \multicolumn{1}{c}{\textbf{Technical Summary}} \\
    \midrule
    \multirow{23}{*}{\rotatebox{90}{\textbf{Prompt-based Methods}}}
    & Ner4OPT \cite{dakle2023ner4opt} & CPAIOR, 2023 & Two-stage & Fine-tune models for named-entity recognition with established NLP techniques. \\
    & AOMG \cite{almonacid2023towards} & GECCO, 2023 & Direct & Use LLMs to generate mathematical optimization models directly. \\
    & HG 2.0 \cite{tsouros2023holy} & arXiv, 2023 & Two-stage & Embed LLMs in a two-stage optimization-modeling framework. \\
    & AMGPT \cite{li2023synthesizing} & arXiv, 2023 & Two-stage & Use fine-tuned models to classify constraints before model generation. \\
    & CoE \cite{xiao2023chain} & ICLR, 2023 & Multi-agent & Construct dynamic reasoning chains with 11 specialized agents. \\
    & OptiMUS \cite{ahmaditeshnizi2024optimus} & ICML, 2023 & Multi-agent & Coordinate modeling, programming, and evaluation through a conductor agent. \\
    & MAMS \cite{mostajabdaveh2024optimization} & INFOR, 2024 & Multi-agent & Use inter-agent cross-validation instead of solver-dependent verification. \\
    & EC \cite{jin2024democratizing} & SGC, 2024 & Two-stage & Apply a two-stage modeling framework to energy-management systems. \\
    & CAFA \cite{deng2024cafa} & NeurIPS, 2024 & Two-stage & Improve model generation through code-based formalization. \\
    & MAMO \cite{huang2024llms} & NAACL, 2024 & Two-stage & Extend optimization-modeling evaluation to ordinary differential equations. \\
    & NL2OR \cite{li2025abstract} & arXiv, 2024 & Two-stage & Constrain LLM outputs with predefined abstract model structures. \\
    & TRIP-PAL \cite{de2024trip} & arXiv, 2024 & Two-stage & Apply a two-stage modeling framework to travel planning. \\
    & OptLLM \cite{zhang2024solving} & arXiv, 2024 & Interactive & Support both single-turn and interactive input modes. \\
    & OptiMUS-0.3 \cite{ahmaditeshnizi2024optimus03} & arXiv, 2024 & Multi-agent & Add self-correction and structure-aware modeling to OptiMUS. \\
    & MeetMate \cite{lawless2024want} & TiiS, 2024 & Interactive & Process user input interactively through five selectable task modes. \\
    & OptiBench \cite{wang2024optibench} & arXiv, 2024 & Two-stage & Verify model equivalence with a modified Weisfeiler-Lehman graph-isomorphism procedure. \\
    & VL \cite{gomez2025values} & CUI, 2025 & Interactive & Translate user priorities into optimization constraints through dialogue. \\
    & LLM-MCTS \cite{astorga2024autoformulation} & arXiv, 2025 & Two-stage & Search the formulation hypothesis space with hierarchical Monte Carlo tree search. \\
    & EquivaMap \cite{zhai2025equivamap} & arXiv, 2025 & Two-stage & Generate variable-mapping functions with LLMs and lightweight verification. \\
    & MAP \cite{talebi2025large} & arXiv, 2025 & Multi-agent & Use multiple independent reviewer agents to assess generated models. \\
    & ORMind \cite{wang2025ormind} & arXiv, 2025 & Multi-agent & Replace conductor-based coordination with structured and predictable workflows. \\
    & LEAN-LLM-OPT \cite{liang2025llm} & SSRN, 2025 & Multi-agent & Use RAG for problem classification and construct instances through query operations. \\
    & OptiTree \cite{liu2026optitree} & NeurIPS, 2026 & Multi-agent & Decompose problems hierarchically in a tree and synthesize the resulting subproblem analyses. \\
    \midrule
    \multirow{14}{*}{\rotatebox{90}{\textbf{Learning-based Methods}}}
    & LM4OPT \cite{ahmed2024lm4opt} & INFOR, 2024 & Fine-tuning & Progressively fine-tune models on the NL4OPT dataset. \\
    & ORLM \cite{huang2025orlm} & OR, 2024 & Data synthesis & Expand and augment training data before fine-tuning open-source models. \\
    & ReSocratic \cite{yang2024optibench} & ICLR, 2024 & Data synthesis & Introduce inverse data synthesis and construct the OPTIBENCH benchmark. \\
    & LLMOPT \cite{jiang2024llmopt} & ICLR, 2024 & Fine-tuning & Combine model alignment with self-correction to reduce hallucinated formulations. \\
    & BPP-Search \cite{wang2024bpp} & arXiv, 2024 & Data synthesis & Recover missing intermediate details during synthetic-data generation. \\
    & OptMATH \cite{lu2025optmath} & arXiv, 2025 & Data synthesis & Develop a scalable bidirectional data-synthesis pipeline. \\
    & LLMBO \cite{amarasinghe2023ai} & arXiv, 2025 & Data synthesis & Fine-tune cost-efficient LLMs for domain-specific business optimization tasks.\\
    & SIRL \cite{chen2025solver} & arXiv, 2025 & Fine-tuning & Use external optimization solvers as verifiable reward evaluators for reinforcement learning. \\
    & Step-Opt \cite{wu2025step} & arXiv, 2025 & Data synthesis & Increase problem complexity through iterative problem generation. \\
    & DPLM \cite{zhou2025auto} & arXiv, 2025 & Data synthesis & Combine the diversity of forward generation with the reliability of inverse generation. \\
    & OptiTrust \cite{lima2025toward} & arXiv, 2025 & Data synthesis & Construct a verifiable pipeline for synthetic-data generation. \\
    & StepORLM \cite{zhou2025steporlm} & arXiv, 2025 & Fine-tuning & Refine reasoning trajectories through generative process supervision and co-evolution. \\
    & OR-R1 \cite{ding2026or} & AAAI, 2026 & Fine-tuning & Combine fine-tuning with TGRPO to improve efficiency and output consistency on unlabeled data. \\
    & DeepOR \cite{xiao2026deepor} & AAAI, 2026 & Fine-tuning & Develop a reasoning foundation model that generates structured optimization formulations.\\
    
    \bottomrule
    \end{tabular}
    \end{adjustbox}
\end{table*}

\newpage

\begin{table*}[t!]
    \centering
    \caption{Representative studies on LLMs for optimization solving, grouped into LLMs as direct optimizers, low-level assistants, and high-level algorithm generators.}
    \label{tab:solving}
    \renewcommand{\arraystretch}{1.2}
    \begin{adjustbox}{max width=\linewidth}
    \begin{tabular}{c|c|c|c|l}
    \toprule
    \textbf{} & \textbf{Method} & \textbf{Venue} & \textbf{Type} & \multicolumn{1}{c}{\textbf{Technical Summary}} \\
    \midrule
    \multirow{16}{*}{\rotatebox{90}{\textbf{LLMs as Optimizers}}}
    & OPRO \cite{yang2023large} & ICLR, 2023 & Prompt-based & Refine candidate solutions iteratively from problem descriptions and optimization trajectories. \\
    & toLLM \cite{guo2023towards} & KDD, 2023 & Prompt-based & Define four canonical tasks to evaluate the optimization limits of LLMs. \\
    & EvoLLM \cite{lange2024large} & GECCO, 2024 & Prompt-based & Replace raw optimization trajectories with rankings of candidate quality. \\
    & MLLMO \cite{huang2025multimodal} & MCII, 2024 & Prompt-based & Use multimodal LLMs to process problem descriptions and route visualizations jointly for CVRPs. \\
    & POM \cite{li2024pretrained} & NeurIPS, 2024 & Learning-based & Pretrain a general-purpose foundation model for zero-shot black-box optimization. \\
    & OPTO \cite{cheng2024trace} & NeurIPS, 2024 & Prompt-based & Replace conventional optimization trajectories with rich execution traces. \\
    & VRMA \cite{elhenawy2024visual} & arXiv, 2024 & Prompt-based & Use multimodal LLMs to process two-dimensional point-distribution maps. \\
    & ROPRO \cite{zhang2024revisiting} & ACL, 2024 & Prompt-based & Analyze OPRO's model dependence and its limitations on smaller models. \\
    & BBOLLM \cite{huang2024exploring} & arXiv, 2024 & Prompt-based & Evaluate LLMs on discrete and continuous black-box optimization tasks. \\
    & LLMS \cite{abgaryan2024llms} & arXiv, 2024 & Learning-based & Fine-tune LLMs on instruction-solution pairs for scheduling problems. \\
    & ECLIPSE \cite{jiang2024optimizable} & NAACL, 2024 & Prompt-based & Apply iterative optimization to the design of jailbreak attacks. \\
    & LLOME \cite{chen2024generalists} & arXiv, 2024 & Learning-based & Use preference learning to satisfy complex biophysical constraints. \\
    & LLMDSM \cite{jiang2024large} & arXiv, 2024 & Prompt-based & Apply iterative optimization to design-structure-matrix sequencing. \\
    & LMCO \cite{qiu2024large} & WCL, 2024 & Prompt-based & Apply iterative optimization to wireless-network design. \\
    & MGSCO \cite{zhao2025bridging} & arXiv, 2025 & Prompt-based & Use multimodal LLMs to interpret visual representations of abstract graphs. \\
    & ORFS \cite{ghose2025orfs} & arXiv, 2025 & Prompt-based & Apply iterative optimization to automated parameter tuning in chip design. \\
    \midrule
    \multirow{25}{*}{\rotatebox{90}{\textbf{Low-level LLMs for Optimization Algorithms}}}
    & LMX \cite{meyerson2024language} & TELO, 2023 & Operators & Use LLMs as operators for crossover and recombination of textual genomes. \\
    & GPT-NAS \cite{yu2023gpt} & arXiv, 2023 & Initialization & Use prior knowledge from LLMs to initialize neural architecture search. \\
    & LMEA \cite{liu2024large} & CEC, 2023 & Operators & Use LLMs as crossover, mutation, and selection operators in EAs. \\
    & LLM-PP \cite{jawahar2023llm} & arXiv, 2023 & Initialization & Use LLM-based performance prediction to support initialization. \\
    & LMOEA \cite{liu2025large} & EMO, 2023 & Operators & Use zero-shot LLM prompting as a search operator within MOEA/D. \\
    & OFPLLM \cite{de2023optimized} & AI, 2023 & Initialization & Assist non-expert users in initializing financial plans. \\
    & LEO \cite{brahmachary2025large} & Neucom, 2024 & Operators & Guide candidates from separate exploration and exploitation pools with LLMs. \\
    & LLM-GA \cite{teukam2024integrating} & BiB, 2024 & Initialization & Use LLMs to initialize high-quality mutant pools for enzyme-design GAs.\\
    & GE \cite{morris2024llm} & GECCO, 2024 & Operators & Use role-based prompts to increase creativity and diversity in LLM-assisted NAS. \\
    & LLMAES \cite{wang2024largeicic} & ICIC, 2024 & Operators & Generate selected population candidates with LLMs to reduce interaction costs. \\
    & LLMTES \cite{kramer2024large} & arXiv, 2024 & Configuration & Tune evolution strategies sequentially through LLM-based feedback. \\
    & LAEA \cite{hao2024large} & SWEVO, 2024 & Evaluation & Reformulate model-assisted selection as classification and regression. \\
    & LICO \cite{nguyen2024lico} & arXiv, 2024 & Evaluation & Use LLMs as surrogate models for molecular-science applications. \\
    & LLMSM \cite{rios2024large} & CAI, 2024 & Evaluation & Coordinate model selection and training for engineering optimization with multiple LLMs. \\
    & LLMCES \cite{custode2024investigation} & GECCO, 2024 & Configuration & Control evolution-strategy step sizes with an OPRO-like mechanism. \\
    & AKFLLM \cite{huang2024advancing} & arXiv, 2024 & Operators & Support mutation and other generative stages in evolutionary multitask optimization with LLMs. \\
    & LLMAMO \cite{liu2024largemulti} & arXiv, 2024 & Operators & Invoke LLMs to generate elite solutions when population improvement is insufficient. \\
    & LTC \cite{kramer2024llama} & ESANN, 2024 & Configuration & Control CMA-ES dynamically through sequential LLM feedback. \\
    & LLMEVO \cite{zhao2025can} & arXiv, 2025 & Operators & Evaluate LLMs in selection, crossover, and mutation, while identifying limitations in initialization. \\
    & LLM-GE \cite{yu2025llm} & GECCO, 2025 & Operators & Use LLMs as crossover and mutation operators for YOLO architecture optimization. \\
    & LAOS \cite{zhang2025laos} & GECCO, 2025 & Configuration & Replace optimization trajectories with state features for LLM-based operator selection. \\
    & PAIR \cite{ali2025pair} & arXiv, 2025 & Operators & Use LLMs primarily as selection operators to extend LMEA. \\
    & LLMMS \cite{zhang2025large} & arXiv, 2025 & Evaluation & Use LLM meta-surrogates and token-sequence representations for cross-task knowledge transfer. \\
    & LMPSO \cite{shinohara2025large} & arXiv, 2025 & Operators & Simulate PSO dynamics with LLMs and adapt LMEA to specific algorithms. \\
    & LLM-SAEA \cite{xie2025large} & arXiv, 2025 & Evaluation & Select surrogate models and infill criteria dynamically with LLMs. \\
    \midrule
    \multirow{30}{*}{\rotatebox{90}{\textbf{High-level LLMs for Optimization Algorithms}}}
    & LLMGM \cite{pluhacek2023leveraging} & GECCO, 2023 & Single-step & Decompose and recombine six metaheuristics to generate hybrid algorithms in a single step. \\
    & FunSearch \cite{romera2024mathematical} & Nature, 2023 & Iterative & Generate code fragments with LLMs and use EAs to search the resulting function space. \\
    & AEL \cite{liu2023algorithm} & arXiv, 2023 & Iterative & Extend FunSearch with heuristic principles and improve performance on TSP instances. \\
    & AS-LLM \cite{wu2023large} & IJCAI, 2023 & Selection & Extract high-dimensional algorithm features from code and text for algorithm selection. \\
    & EoH \cite{liu2024evolution} & ICML, 2024 & Iterative & Co-evolve heuristic descriptions and code through predefined crossover and mutation prompts. \\
    & ReEvo \cite{ye2024reevo} & NeurIPS, 2024 & Iterative & Guide hyper-heuristic search through reflective evolution with LLMs.\\
    & AutoSAT \cite{sun2024autosat} & arXiv, 2024 & Iterative & Combine multiple heuristic strategies to guide LLM-based algorithm generation. \\
    & LLaMEA \cite{van2024llamea} & TEVC, 2024 & Iterative & Use LLMs within evolutionary mutation and selection to generate heuristics. \\
    & L-AutoDA \cite{guo2024autoda} & GECCO, 2024 & Iterative & Apply AEL to generate adversarial-attack algorithms for cybersecurity. \\
    & MOELLM \cite{huang2025autonomous} & TEVC, 2024 & Iterative & Combine robust testing with dynamic selection to generate multi-objective optimization algorithms. \\
    & LLMEPS \cite{zhang2024understanding} & PPSN, 2024 & Iterative & Establish a baseline for automated algorithm design based on EoH and ReEvo. \\
    & EvolCAF \cite{yao2024evolve} & PPSN, 2024 & Iterative & Apply EoH to evolve cost-aware acquisition functions for Bayesian optimization. \\
    & TS-EoH \cite{yatong2024ts} & arXiv, 2024 & Iterative & Apply EoH to algorithm generation for edge-server task scheduling. \\
    & MEoH \cite{yao2025multi} & AAAI, 2024 & Iterative & Extend EoH to multi-objective search with additional algorithm-quality criteria. \\
    & LLaMEA-HPO \cite{van2024loop} & TELO, 2024 & Iterative & Integrate hyperparameter optimization into the LLaMEA search cycle. \\
    & CMLLM \cite{yin2025controlling} & EvoApps, 2024 & Iterative & Control LLM mutation through dynamic prompts to improve LLaMEA. \\
    & HSEvo \cite{dat2025hsevo} & AAAI, 2024 & Iterative & Combine harmony search with GAs and optimize portfolio quality and diversity. \\
    & LLM4AD \cite{liu2024llm4ad} & arXiv, 2024 & Iterative & Provide an EoH-based platform for heuristic algorithm design. \\
    & MCTS-AHD \cite{zheng2025monte} & arXiv, 2025 & Iterative & Organize generated heuristics in a tree and explore them with MCTS. \\
    & CEG \cite{van2025code} & GECCO, 2025 & Iterative & Analyze LLM-generated code and its evolutionary dynamics during search. \\
    & OPSLAD \cite{yin2025optimizing} & GECCO, 2025 & Iterative & Apply LLaMEA to industrial photonic-structure optimization. \\
    & BLADE \cite{van2025blade} & GECCO, 2025 & Iterative & Establish a standardized and reproducible evaluation framework for LLM-driven algorithm generation (LLaMEA specifically). \\
    & AutoHD \cite{ling2025complex} & arXiv, 2025 & Iterative & Discover heuristics for complex planning tasks under LLM guidance. \\
    & LAMA \cite{li2025llm} & TEVC, 2025 & Iterative & Construct an automated memetic algorithm with LLM-designed heuristics. \\
    & LAS \cite{liu2025fitness} & arXiv, 2025 & Iterative & Analyze the fitness landscapes of LLM-assisted algorithm search. \\
    & LLaMEA-BO \cite{li2025llamea} & arXiv, 2025 & Iterative & Apply LLaMEA to automated Bayesian-optimization algorithm design. \\
    & InstSpecHH \cite{zhang2025llm} & arXiv, 2025 & Selection & Use LLM semantic reasoning for context-aware algorithm selection. \\
    & BSALMD \cite{van2025behaviour} & arXiv, 2025 & Iterative & Analyze behavioral spaces to characterize algorithm-evolution trajectories. \\
    & FLAAD \cite{liu2025fine} & arXiv, 2025 & Iterative & Fine-tune LLMs for algorithm generation with diversity-aware ranked sampling. \\
    & EoH-S \cite{liu2025eoh} & arXiv, 2025 & Iterative & Extend EoH to evolve complementary portfolios of algorithms. \\
    \bottomrule
    \end{tabular}
    \end{adjustbox}
\end{table*}

\section{Representative Benchmarks \label{sec:benchmark}}

As discussed in the main text, an effective benchmark serves two purposes. It provides a diverse problem set that exposes the strengths and limitations of competing methods, and it supports centralized evaluation under comparable conditions. Table~\ref{tab:benchmark} summarizes representative benchmarks for optimization modeling and solving. For each benchmark, we report the venue, scale, problem scope, and construction strategy. The modeling benchmarks progress from competition-derived datasets and manual curation to scalable synthetic generation. The solving benchmarks range from single-turn graph and path-planning tasks to agentic suites for iterative algorithm design. Together, they reflect the field's transition from direct problem solving to the generation and evaluation of optimization algorithms.

\begin{table*}[h]
\centering
\caption{Representative benchmarks for LLM-driven optimization modeling and solving.}
\label{tab:benchmark}
\renewcommand{\arraystretch}{1.2}

\begin{adjustbox}{max width=\linewidth}
\begin{tabular}{c|c|c|c|l}
\toprule
 & \textbf{Benchmark} & \textbf{Venue} & \textbf{Scale} & \multicolumn{1}{c}{\textbf{Technical Summary}} \\
\midrule

\multirow{10}{*}{\rotatebox{90}{\textbf{Modeling}}}
& NL4Opt \cite{ramamonjison2023nl4opt} & EMNLP, 2022 & 245 & Competition-derived benchmark for translating natural language into LP models. \\
& NLP4LP \cite{ahmaditeshnizi2024optimus03} & ICML, 2024 & 67 & Natural-language descriptions of LP and MILP models. \\
& NL2OPT \cite{mostajabdaveh2024optimization} & INFOR, 2024 & 70 & Handcrafted LP, MILP, and QP instances with structured specifications and Zimpl ground truth. \\
& ComplexOR \cite{xiao2023chain} & ICLR, 2024 & 37 & Complex MILP problems that require multistep formulation. \\
& MAMO \cite{huang2024llms} & NAACL, 2025 & 1,209 & Two difficulty levels spanning LP, MILP, and ordinary differential equations. \\
& IndustryOR \cite{huang2025orlm} & OR, 2025 & 100 & Synthetically generated industrial problems covering LP, IP, MILP, and NLP. \\
& OptiBench \cite{yang2024optibench} & ICLR, 2025 & 605 & Synthetic LP, IP, MILP, and NLP instances with controllable complexity and verified equivalence. \\
& DP-Bench \cite{zhou2025auto} & arXiv, 2025 & 132 & Textbook-derived deterministic and stochastic dynamic-programming problems with finite or infinite horizons. \\
& DCP-Bench \cite{michailidis2025cp} & ECAI, 2025 & 164 & Natural-language CSP and COP instances with constraint-programming ground truth. \\
& OptMath \cite{lu2025optmath} & ICML, 2025 & 165 & Scalable bidirectional synthesis across LP, IP, MILP, NLP, and SOCP formulations. \\
\midrule

\multirow{9}{*}{\rotatebox{90}{\textbf{Solving}}}
& NLGraph \cite{wang2023can} & NeurIPS, 2023 & 29,370 & Eight polynomial-time graph-reasoning tasks for systematic LLM evaluation. \\
& PPNL \cite{aghzal2023can} & ICLR, 2024 & 160,000+ & Controllable grid-world path-planning tasks for spatial and temporal reasoning. \\
& GraphArena \cite{tang2025grapharena} & ICLR, 2025 & 10,000 & Real-world graph benchmark covering both P-class and NP-complete tasks. \\
& ALE-Bench \cite{imajuku2026ale} & NeurIPS, 2025 & -- & Long-horizon, objective-driven algorithm engineering with Elo-style scoring. \\
& OPT-BENCH \cite{li2025opt} & arXiv, 2025 & 30 & Large-search-space benchmark with an OPT-Agent for Kaggle ML tasks and NP problems. \\
& CO-Bench \cite{sun2026co} & AAAI, 2026 & 6,482 & Benchmark for LLM-based algorithm search across 36 real-world CO problems in eight categories. \\
& FrontierCO \cite{feng2026frontierco} & ICLR, 2026 & 258+ & Competition-scale TSP, CVRP, and MIS instances with easy and hard splits. \\
& HeuriGym \cite{chen2025heurigym} & ICLR, 2026 & 9 & Low-exposure CO problems for evaluating LLM-generated heuristics with the Quality-Yield Index. \\
& LLM4AD \cite{liu2024llm4ad} & arXiv, 2026 & 160+ & Unified platform for LLM-based algorithm design with modular search, sandboxing, and a GUI. \\
\bottomrule

\end{tabular}
\end{adjustbox}
\end{table*}

\section{Detailed Evaluation of Optimization Modeling Methods \label{sec:modeling}}

To support the empirical synthesis in the main text, we evaluate a broad set of baselines for optimization modeling. Table~\ref{tab:modeling_performance} reports the complete performance matrix. The baseline category includes the general-purpose models GPT-3.5-Turbo, GPT-4, GPT-4o, and DeepSeek-V3, together with the reasoning models DeepSeek-R1, OpenAI-o1, and OpenAI-o3. The prompt-based category includes CoE \cite{xiao2023chain}, CoT \cite{wei2022chain}, CAFA \cite{deng2024cafa}, OptiMUS \cite{ahmaditeshnizi2024optimus}, LLM-MCTS \cite{astorga2024autoformulation}, StepORLM+GenPRM \cite{zhou2025steporlm}, and OptiTree \cite{liu2026optitree}. The learning-based category includes ORLM \cite{huang2025orlm}, OptMATH \cite{lu2025optmath}, LLMOPT \cite{jiang2024llmopt}, StepORLM \cite{zhou2025steporlm}, OptiTrust \cite{lima2025toward}, Step-Opt \cite{wu2025step}, SIRL \cite{chen2025solver}, DeepOR \cite{xiao2026deepor}, and OR-R1 \cite{ding2026or}. The main text aggregates 11 representative methods for the average-accuracy analysis, whereas this supplementary document reports all available results at the dataset level.

We collect the results through a three-stage protocol. First, we extract metrics reported in the original publications. Second, we incorporate values from official benchmark reports after cross-checking the method and metric definitions. Third, we fill remaining gaps with results from credible reproduction studies, including the reproductions reported by OptiTree. The evaluation environments are not fully homogeneous. They differ in solver configurations, few-shot examples, decoding temperatures, and the number of trials used to estimate pass@1. Consequently, cross-paper comparisons require caution. We reduce this risk by prioritizing original and benchmark-reported values and by aligning metric definitions whenever possible.

\begin{figure*}[t!]
    \centering
    \begin{minipage}[t]{0.62\textwidth}
        \centering
        \includegraphics[width=\linewidth]{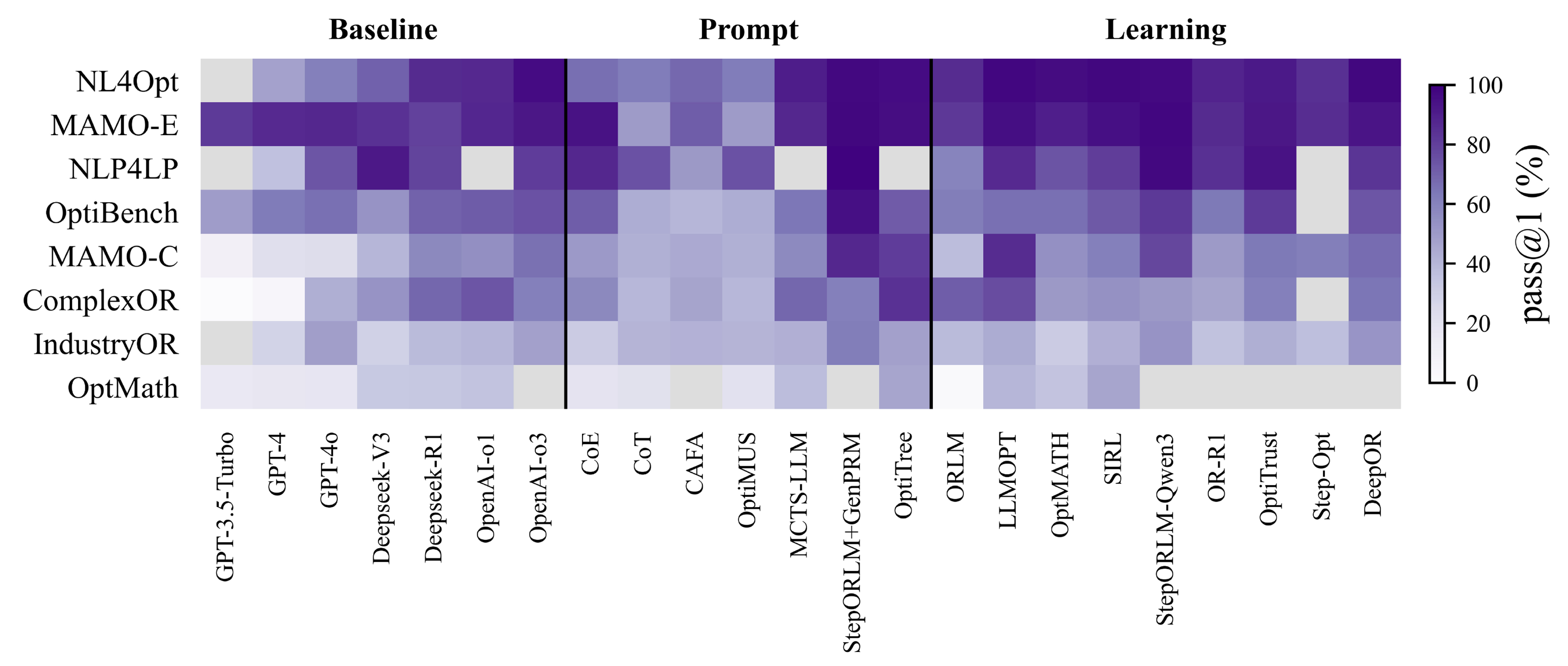}
        \vspace{1pt}
        \centerline{\small (a) Performance heatmap}
    \end{minipage}
    \hfill
    \begin{minipage}[t]{0.35\textwidth}
        \centering
        \includegraphics[width=\linewidth]{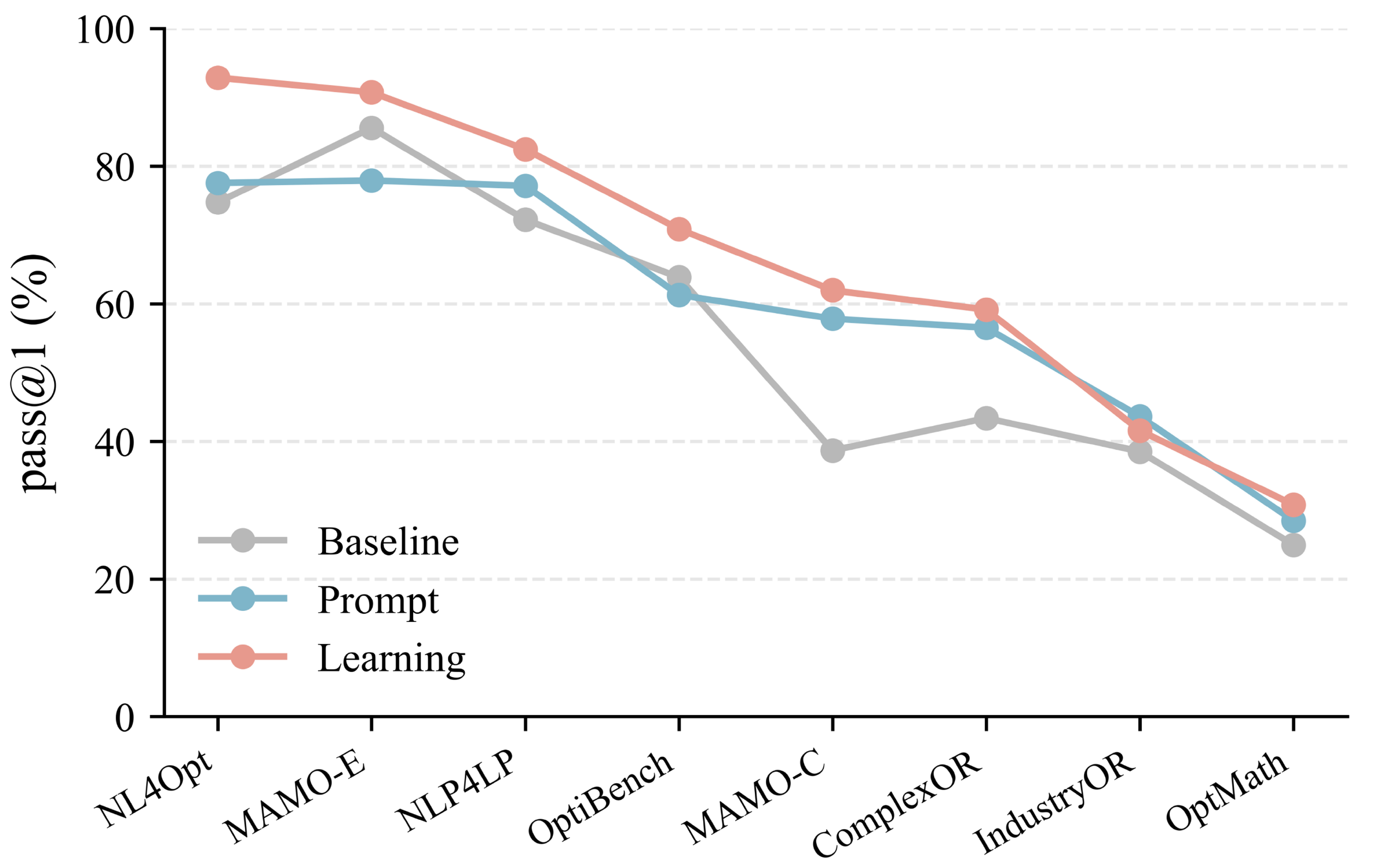}
        \vspace{1pt}
        \centerline{\small (b) Category-mean trajectories}
    \end{minipage}
    \caption{Detailed performance on eight optimization-modeling benchmarks. (a)~Per-dataset pass@1 for all 23 methods, with datasets ordered by increasing difficulty; gray cells indicate unreported results. (b)~Category-mean pass@1 over the same dataset order; the curves converge on easier datasets and separate as difficulty increases.}
    \label{fig:modeling_appendix}
    \vspace{-1em}
\end{figure*}

Table~\ref{tab:modeling_performance} reports results for all 23 methods, while Fig.~\ref{fig:modeling_appendix}(a) and (b) visualize the same data as a performance heatmap and category-level trajectories. The overall ordering is consistent with the main-text analysis. Learning-based methods perform best on average, frontier reasoning models follow closely, and prompt-based methods show the greatest internal variation. On easier datasets, such as NL4Opt and MAMO-E, all three categories approach saturation and their performance differences remain small. The categories separate on moderately difficult datasets, including NLP4LP and OptiBench, where stronger prompt-based and learning-based methods outperform weaker baselines. The largest gaps appear on MAMO-C, ComplexOR, IndustryOR, and OptMath. On these datasets, weaker baselines deteriorate sharply, whereas learning-based and search-augmented prompt-based methods retain substantially higher accuracy. Strong reasoning models, particularly DeepSeek-R1, OpenAI-o1, and OpenAI-o3, remain competitive on the difficult datasets and narrow the advantage of specialized learning-based methods. This trend also challenges conventional prompt orchestration. As the reasoning capabilities of base models improve, future gains are more likely to come from structured multi-agent interaction and semantic-space search than from standard prompting alone.

A second finding concerns evaluation coverage. Many entries in Table~\ref{tab:modeling_performance} are unreported, and the missing values are concentrated on the most difficult datasets and among the strongest methods. These methods are often evaluated only on training-aligned or author-selected subsets. Consequently, headline scores are not directly comparable, and strong results on partial subsets may not generalize. Future benchmark research should address this limitation in two ways. First, the community needs an agentic optimization-modeling platform that evaluates common prompt-based and learning-based methods under a shared protocol. Such a platform would enable direct comparison and independent validation of partial-coverage claims. Second, prompt-based methods should be re-evaluated with newer and stronger base models. Most existing studies use GPT-4o or earlier models, so evaluations on frontier backbones are necessary to determine the remaining contribution of prompt-level strategies.

\begin{table*}[t!]
    \centering
    \caption{Performance on optimization-modeling benchmarks. All values are accuracy (\%). $^*$ denotes values reported in the original papers, $^\dagger$ denotes values reported in benchmark studies, and unmarked values come from credible reproduction studies. The best result within each method category is boldfaced.}
    \label{tab:modeling_performance}
    \renewcommand{\arraystretch}{1.2}
    \begin{adjustbox}{max width=\linewidth}
    \begin{tabular}{c|c|c|c|c|c|c|c|c|c|c}
    \toprule
    & \textbf{Method} & \textbf{NL4Opt} & \textbf{MAMO-E} & \textbf{MAMO-C} & \textbf{NLP4LP} & \textbf{ComplexOR} & \textbf{IndustryOR} & \textbf{OptiBench} & \textbf{OptMath} & \textbf{Avg.} \\
    \midrule
    \multirow{7}{*}{\rotatebox{90}{\textbf{Baseline}}}
    & GPT-3.5-Turbo & -- & 81.3$^\dagger$ & 9.5$^\dagger$ & -- & 0.5$^\dagger$ & -- & 49.1$^\dagger$ & 15.0$^\dagger$ & -- \\
    & GPT-4 & 47.3 & 86.5$^\dagger$ & 21.1$^\dagger$ & 35.8$^\dagger$ & 4.9$^\dagger$ & 28.0$^\dagger$ & 62.8$^\dagger$ & 16.6$^\dagger$ & 37.9 \\
    & GPT-4o & 61.2 & 87.3$^\dagger$ & 22.8$^\dagger$ & 73.6 & 42.9 & 48.4 & 66.1$^\dagger$ & 17.5 & 52.5 \\
    & DeepSeek-V3 & 70.5 & 84.3 & 39.8 & \textbf{92.1} & 52.6 & 29.0 & 52.4 & 32.6$^\dagger$ & 56.7 \\
    & DeepSeek-R1 & 86.1 & 79.5 & 57.3 & 78.6 & 68.4 & 38.0 & 70.2 & 33.1 & \textbf{63.9} \\
    & OpenAI-o1 & 87.1 & 87.6 & 54.5 & -- & \textbf{73.6} & 40.0 & 71.5 & \textbf{34.9} & -- \\
    & OpenAI-o3 & \textbf{96.2} & \textbf{92.4} & \textbf{65.8} & 81.0 & 60.7 & \textbf{47.8} & \textbf{74.8} & -- & -- \\
    \midrule
    \multirow{7}{*}{\rotatebox{90}{\textbf{Prompt-based}}}
    & Chain-of-Experts & 66.7 & 94.4 & 50.6 & 87.4 & 57.1 & 31.2 & 71.2 & 18.6 & \textbf{59.7} \\
    & Chain-of-Thought & 62.2 & 49.5 & 42.3 & 74.7 & 39.2 & 40.5 & 43.6 & 20.5 & 46.6 \\
    & CAFA & 68.1 & 71.2 & 44.5 & 50.0 & 46.4 & 41.1 & 40.1 & -- & -- \\
    & OptiMUS & 62.2 & 49.5 & 42.3 & 74.7 & 39.2 & 40.5 & 43.6 & 20.2$^\dagger$ & 46.5 \\
    & MCTS-LLM & 90.3 & 87.4 & 56.8 & -- & 68.4 & 42.0 & 64.0 & 37.3 & -- \\
    & StepORLM+GenPRM & \textbf{97.2}$^*$ & \textbf{97.8}$^*$ & \textbf{87.4}$^*$ & \textbf{98.9}$^*$ & 61.1$^*$ & \textbf{61.9}$^*$ & \textbf{94.6}$^*$ & -- & -- \\
    & OptiTree & 96.2$^*$ & 95.6$^*$ & 81.0$^*$ & -- & \textbf{84.2}$^*$ & 48.0$^*$ & 71.9$^*$ & \textbf{45.8}$^*$ & -- \\
    \midrule
    \multirow{9}{*}{\rotatebox{90}{\textbf{Learning-based}}}
    & ORLM-LLaMA-3-8B & 85.7$^*$ & 82.3$^*$ & 37.4$^*$ & 59.5 & 71.4 & 38.0$^*$ & 61.8 & 2.6$^\dagger$ & 54.8 \\
    & LLMOPT-Qwen2.5-14B & 97.3$^*$ & 95.3$^*$ & \textbf{85.8}$^*$ & 86.5$^*$ & \textbf{76.5}$^*$ & 44.0$^*$ & 66.4$^*$ & 40.0$^*$ & \textbf{74.0} \\
    & OptMATH-Qwen2.5-32B & 95.9$^*$ & 89.9$^*$ & 54.1$^*$ & 73.9 & 50.0 & 31.0$^*$ & 66.1$^*$ & 34.7$^*$ & 62.0 \\
    & SIRL-Qwen2.5-32B & \textbf{98.0}$^*$ & 94.6$^*$ & 61.1$^*$ & 80.6 & 53.6 & 42.0$^*$ & 72.6 & \textbf{45.8}$^*$ & 68.5 \\
    & StepORLM-Qwen3-8B & 96.7$^*$ & \textbf{97.6}$^*$ & 77.5$^*$ & \textbf{97.2}$^*$ & 50.0$^*$ & \textbf{52.4}$^*$ & \textbf{81.9}$^*$ & -- & -- \\
    & OR-R1-Qwen3-8B & 88.3$^*$ & 86.1$^*$ & 49.9$^*$ & 84.6$^*$ & 46.3$^*$ & 35.3$^*$ & 62.9$^*$ & -- & -- \\
    & OptiTrust-Granite3.2-8B & 91.6$^*$ & 92.3$^*$ & 63.1$^*$ & 94.4$^*$ & 61.1$^*$ & 42.9$^*$ & 81.4$^*$ & -- & -- \\
    & Step-Opt-LLaMA-3-8B & 84.5$^*$ & 85.3$^*$ & 61.6$^*$ & -- & -- & 36.4$^*$ & -- & -- & -- \\
    & DeepOR-Qwen3-8B & 97.7$^*$ & 93.2$^*$ & 67.1$^*$ & 82.9$^*$ & 64.3$^*$ & 52.2$^*$ & 73.8$^*$ & -- & -- \\
    \bottomrule
    \end{tabular}
    \end{adjustbox}
\end{table*}

\section{Detailed Evaluation of LLMs as Optimizers \label{sec:Detailed Evaluation of LLM-as-Optimizer}}

To support the empirical claims in the main text, we report complete per-function results for the LLM-as-optimizer study in Table~\ref{tab:optimizer_full}. The classical group contains four established black-box optimizers: DE, PSO, CMA-ES, and random search. All four are tensorized in the EvoX framework \cite{huang2024evox}. The LLM group follows the OPRO paradigm and iteratively generates raw floating-point candidate vectors at a sampling temperature of 1.0. We evaluate GPT-5-mini, DeepSeek-V4, and Qwen3.5. Chain-of-thought reasoning is disabled because the task requires direct numerical candidate generation rather than extended verbal reasoning. GPT-5-mini uses its minimum reasoning-effort setting, while the explicit thinking modes of DeepSeek-V4 and Qwen3.5 are disabled.

Both groups receive the same evaluation budget. Each method evaluates eight candidates per iteration for 200 iterations, in addition to the eight initial candidates, for a total of 1608 function evaluations per run. The test suite contains four BBOB functions \cite{hansen2021coco} with distinct landscape properties: Sphere, Rosenbrock, Discus, and Schwefel. We evaluate dimensions 2, 5, and 10 with five independent random seeds for each configuration. Performance is measured by the relative improvement $r$, defined as

\begin{equation}
r = \operatorname{clip}_{[0,1]} \left( \frac{f_{\mathrm{init}} - f_{\mathrm{best}}}{f_{\mathrm{init}} - f_{\mathrm{opt}}} \right),
\end{equation}
where $f_{\mathrm{init}}$ is the best fitness among the initial random samples, $f_{\mathrm{best}}$ is the best fitness obtained by the end of the run, and $f_{\mathrm{opt}}$ is the known optimum of the corresponding BBOB instance. We use this scaled metric because the benchmark functions operate on markedly different numerical scales. Typical values are on the order of $10^3$ for Sphere and $10^5$ for Discus, and the instance optima also vary across dimensions. Raw fitness values are therefore not comparable across functions or dimensions. In contrast, $r$ measures the fraction of the initial-to-optimal gap closed by an optimizer. The normalization removes both function-specific scale and dimension-specific optimum effects, which permits the cross-function and cross-dimension averages reported in the main text. Clipping the metric to $[0,1]$ assigns zero improvement to runs that do not improve on the initialization and prevents minor numerical overshoots beyond the known optimum from distorting the averages.

\begin{table*}[t]
\centering
\caption{Per-function relative improvement $r$ (mean $\pm$ standard deviation over five seeds) on four BBOB functions at dimensions 2, 5, and 10. Classical EAs and LLM optimizers use the same budget of 1608 function evaluations. The best value in each method category, classical or LLM-based, is boldfaced in each column.}
\label{tab:optimizer_full}
\renewcommand{\arraystretch}{1.2}
\begin{adjustbox}{max width=\linewidth}
\begin{tabular}{c|c|c|c|c|c|c}
\toprule
\textbf{Dim} & \textbf{Method} & \textbf{F1 (Sphere)} & \textbf{F8 (Rosenbrock)} & \textbf{F11 (Discus)} & \textbf{F20 (Schwefel)} & \textbf{Avg.} \\
\midrule
\multirow{7}{*}{\textbf{2}} & CMA-ES & \textbf{1.000 $\pm$ 0.000} & \textbf{1.000 $\pm$ 0.000} & \textbf{1.000 $\pm$ 0.000} & \textbf{0.945 $\pm$ 0.075} & \textbf{0.986} \\
 & PSO & \textbf{1.000 $\pm$ 0.000} & 0.999 $\pm$ 0.002 & \textbf{1.000 $\pm$ 0.000} & 0.871 $\pm$ 0.127 & 0.968 \\
 & DE & 0.999 $\pm$ 0.003 & 0.963 $\pm$ 0.068 & \textbf{1.000 $\pm$ 0.000} & 0.895 $\pm$ 0.151 & 0.964 \\
 & Random & 0.859 $\pm$ 0.282 & 0.970 $\pm$ 0.039 & 0.999 $\pm$ 0.001 & 0.782 $\pm$ 0.387 & 0.902 \\
 \cmidrule{2-7}
 & GPT-5-mini & \textbf{0.999 $\pm$ 0.001} & \textbf{0.932 $\pm$ 0.041} & 0.917 $\pm$ 0.186 & 0.736 $\pm$ 0.251 & \textbf{0.896} \\
 & DeepSeek-V4 & 0.916 $\pm$ 0.140 & 0.794 $\pm$ 0.396 & \textbf{1.000 $\pm$ 0.000} & \textbf{0.761 $\pm$ 0.201} & 0.868 \\
 & Qwen3.5 & 0.265 $\pm$ 0.411 & 0.680 $\pm$ 0.400 & 0.913 $\pm$ 0.193 & 0.616 $\pm$ 0.382 & 0.619 \\
\midrule
\multirow{7}{*}{\textbf{5}} & CMA-ES & \textbf{1.000 $\pm$ 0.000} & 0.998 $\pm$ 0.004 & 0.994 $\pm$ 0.012 & \textbf{1.000 $\pm$ 0.001} & \textbf{0.998} \\
 & PSO & \textbf{1.000 $\pm$ 0.000} & \textbf{1.000 $\pm$ 0.000} & \textbf{1.000 $\pm$ 0.000} & 0.985 $\pm$ 0.033 & 0.996 \\
 & DE & 0.887 $\pm$ 0.071 & 0.974 $\pm$ 0.036 & \textbf{1.000 $\pm$ 0.000} & 0.994 $\pm$ 0.012 & 0.964 \\
 & Random & 0.775 $\pm$ 0.141 & 0.870 $\pm$ 0.245 & 0.996 $\pm$ 0.007 & 0.998 $\pm$ 0.002 & 0.910 \\
 \cmidrule{2-7}
 & GPT-5-mini & 0.361 $\pm$ 0.156 & 0.710 $\pm$ 0.224 & 0.786 $\pm$ 0.299 & 0.866 $\pm$ 0.168 & 0.681 \\
 & DeepSeek-V4 & \textbf{0.795 $\pm$ 0.301} & \textbf{0.897 $\pm$ 0.130} & \textbf{0.851 $\pm$ 0.313} & \textbf{0.993 $\pm$ 0.011} & \textbf{0.884} \\
 & Qwen3.5 & 0.258 $\pm$ 0.244 & 0.600 $\pm$ 0.413 & 0.803 $\pm$ 0.440 & 0.673 $\pm$ 0.311 & 0.583 \\
\midrule
\multirow{7}{*}{\textbf{10}} & CMA-ES & \textbf{1.000 $\pm$ 0.000} & \textbf{0.999 $\pm$ 0.001} & 0.991 $\pm$ 0.018 & \textbf{1.000 $\pm$ 0.000} & \textbf{0.997} \\
 & PSO & 0.972 $\pm$ 0.037 & 0.999 $\pm$ 0.001 & 1.000 $\pm$ 0.001 & 0.979 $\pm$ 0.046 & 0.987 \\
 & DE & 0.663 $\pm$ 0.129 & 0.876 $\pm$ 0.149 & \textbf{1.000 $\pm$ 0.000} & 0.991 $\pm$ 0.007 & 0.883 \\
 & Random & 0.648 $\pm$ 0.106 & 0.914 $\pm$ 0.087 & 0.992 $\pm$ 0.012 & 0.903 $\pm$ 0.144 & 0.864 \\
 \cmidrule{2-7}
 & GPT-5-mini & 0.200 $\pm$ 0.159 & 0.444 $\pm$ 0.313 & 0.910 $\pm$ 0.169 & 0.509 $\pm$ 0.070 & 0.516 \\
 & DeepSeek-V4 & \textbf{0.630 $\pm$ 0.164} & \textbf{0.865 $\pm$ 0.221} & \textbf{0.988 $\pm$ 0.026} & \textbf{0.837 $\pm$ 0.177} & \textbf{0.830} \\
 & Qwen3.5 & 0.146 $\pm$ 0.200 & 0.587 $\pm$ 0.413 & 0.802 $\pm$ 0.413 & 0.580 $\pm$ 0.391 & 0.529 \\
\bottomrule
\end{tabular}
\end{adjustbox}
\end{table*}

\begin{table}[t]
\centering
\caption{Computational cost and failure statistics for each LLM in the optimizer study, aggregated over 60 runs per model.}
\label{tab:optimizer_cost}
\renewcommand{\arraystretch}{1.2}
\begin{adjustbox}{max width=\linewidth}
\begin{tabular}{c|c|c|c|c|c|c}
\toprule
\textbf{Method} & \textbf{API calls} & \textbf{Input (M)} & \textbf{Output (M)} & \textbf{Wall-clock/run (s)} & \textbf{Fallback events} & \textbf{Stagnant runs (/60)} \\
\midrule
Classical EAs & -- & -- & -- & $\sim$\,seconds & 0 & 0 \\
GPT-5-mini    & 12000 & 14.10 & 2.91 & 1785 & 11 & 5 \\
DeepSeek-V4   & 12000 & 13.65 & 2.83 & 694  & 56 & 1  \\
Qwen3.5       & 12000 & 18.15 & 3.74 & 882  & 690 & 16 \\
\midrule
LLM total     & 36000 & 45.90 & 9.48 & --  & 757 & 22 \\
\bottomrule
\end{tabular}
\end{adjustbox}
\end{table}

Table~\ref{tab:optimizer_full} reports the relative improvement of all seven methods across the three dimensions. The classical optimizers remain consistently strong. CMA-ES achieves average improvements of 0.986, 0.998, and 0.997 at dimensions 2, 5, and 10, respectively, while every classical method remains above 0.86 on average at each dimension. The LLM optimizers perform substantially worse overall, and their performance generally declines as the dimension increases. This weakness is visible even on Sphere, which the classical methods solve reliably: at dimension 10, GPT-5-mini and Qwen3.5 attain only 0.200 and 0.146, respectively. Their stronger results on Discus at the same dimension show that landscape structure also affects performance, but increasing dimension remains a major source of degradation. No LLM dominates across all settings. GPT-5-mini achieves the highest LLM average at dimension 2, whereas DeepSeek-V4 leads at dimensions 5 and 10. DeepSeek-V4 also shows a nonmonotonic trend, with averages of 0.868, 0.884, and 0.830 across the three dimensions. These results support the main-text conclusion that current LLM-driven iterative optimization does not match mature evolutionary algorithms on continuous problems and scales poorly with dimension. The pattern suggests that limited numerical-search capability, rather than any single landscape property, is the principal bottleneck.

Table~\ref{tab:optimizer_cost} summarizes computational overhead and model failures. Each LLM run requires 200 sequential API calls. The mean wall-clock time per run ranges from 694 seconds for DeepSeek-V4 to 1785 seconds for GPT-5-mini, whereas a classical optimizer completes a run within a few seconds on a standard local device. Token consumption is also substantial. Across 60 runs per model, the three LLMs consume 45.90 million input tokens and 9.48 million output tokens. Input tokens dominate because OPRO appends the cumulative optimization trajectory to the prompt at every iteration.

\begin{figure*}[t]
    \centering
    \includegraphics[width=0.99\textwidth]{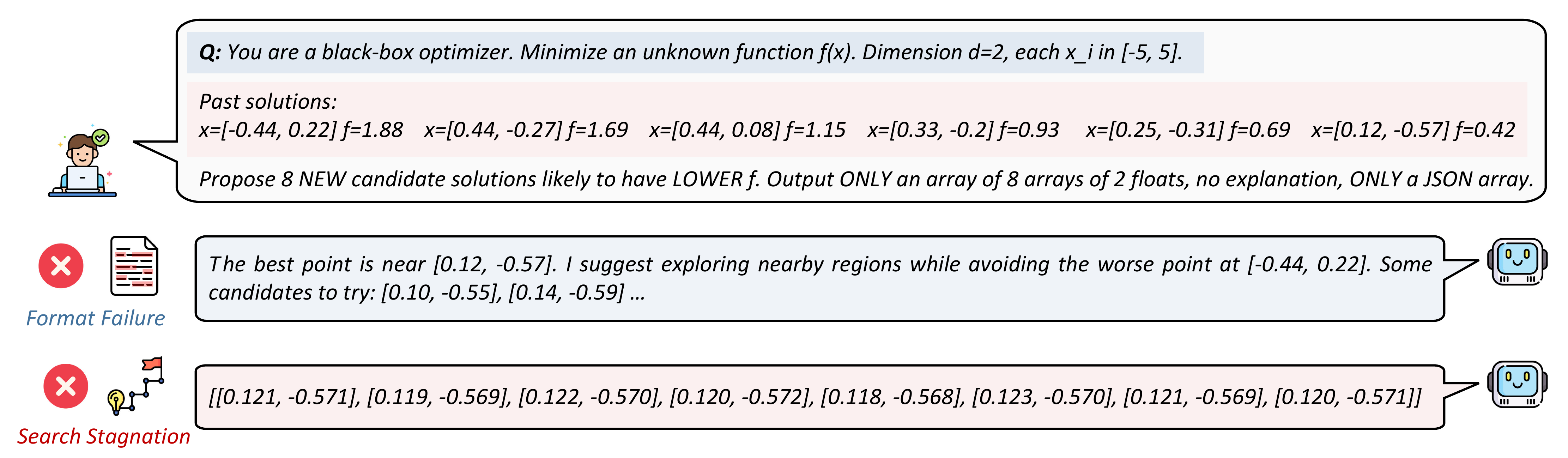}
    \caption{Representative examples of the two main LLM-as-optimizer failure modes: format failure and search stagnation.}
    \label{fig:failures_of_optimizer}
\end{figure*}

Two recurrent failure modes appear during search, and Fig.~\ref{fig:failures_of_optimizer} illustrates representative examples. A format failure occurs when the model output cannot be parsed as the required candidate array. The system then invokes a perturbation-based fallback. Table~\ref{tab:optimizer_cost} records 11 such events for GPT-5-mini, 56 for DeepSeek-V4, and 690 for Qwen3.5. Search stagnation is the second failure mode. It occurs when the model returns valid candidates but fails to improve the objective. We classify a run as stagnant when its final relative improvement satisfies $r<0.2$. Under this criterion, 16 of the 60 Qwen3.5 runs are stagnant, compared with five GPT-5-mini runs and one DeepSeek-V4 run. The clearest example occurs on two-dimensional Sphere, where four of the five Qwen3.5 runs plateau near their initial fitness without triggering a fallback. This behavior is not unique to the OPRO implementation. It indicates a broader tendency of current foundation models to favor conservative local exploitation over effective exploration in continuous floating-point spaces. Section~\ref{Detailed Evaluation of Low-level Assistance} examines the same limitation in the context of dynamic parameter control.

\section{Detailed Evaluation of Low-Level Assistance \label{Detailed Evaluation of Low-level Assistance}}

To support the empirical claims in the main text, we examine dynamic control of the step size in a (1+1)-ES. We compare LLM-based controllers with representative methods from the MetaBBO family \cite{ma2025toward}, which is the most extensively studied family of learned optimizers in this setting. The trained controllers include reinforcement-learning methods \cite{eiben2006reinforcement} and an evolution-strategy-based method \cite{lange2023discovering}. DDQN \cite{sharma2019deep} and PPO \cite{ma2025toward} represent the value-based and policy-based branches of MetaBBO-RL, respectively.

All trained and LLM-based controllers receive the same state representation and reward signal. The state contains the remaining evaluation budget, the number of generations since the last improvement, the relative fitness improvement achieved so far, the recent offspring-acceptance rate, the current step size, the best solution, and its fitness. These descriptors are widely used in dynamic algorithm configuration \cite{cenikj2026survey} and capture information from the optimization process, solution space, and fitness landscape. All controllers choose from the same discrete action space. We conduct the experiments at dimension 20 with a budget of 1000 function evaluations and update the step size every 20 generations.

\begin{table*}[t!]
    \centering
    \caption{Per-function relative improvement $r$ (mean over five seeds) for step-size control in a (1+1)-ES at dimension 20 with a budget of 1000 function evaluations. The controller updates the step size every 20 generations. The best result within each method category is boldfaced.}
    \label{tab:opo_full}
    \renewcommand{\arraystretch}{1.2}
    \begin{tabular}{l|c|c|c|c|c}
    \toprule
    \textbf{Method} & \textbf{Rastrigin} & \textbf{Ackley} & \textbf{Griewank} & \textbf{Schwefel} & \textbf{Avg.} \\
    \midrule
    fixed $\sigma$=0.8 & 0.605 & 0.031 & 0.269 & 0.762 & 0.417 \\
    \midrule
    DDQN        & \textbf{0.626} & \textbf{0.080} & \textbf{0.404} & \textbf{0.658} & \textbf{0.442} \\
    PPO         & 0.467 & 0.021 & 0.312 & 0.572 & 0.343 \\
    OpenAI-ES   & 0.225 & 0.039 & 0.144 & 0.437 & 0.211 \\
    \midrule
    GPT-5-mini  & 0.607 & 0.039 & 0.560 & 0.774 & 0.495 \\
    DeepSeek-V4 & 0.640 & \textbf{0.043} & \textbf{0.596} & 0.731 & 0.503 \\
    Qwen3.5     & \textbf{0.658} & 0.041 & 0.574 & \textbf{0.771} & \textbf{0.511} \\
    \bottomrule
    \end{tabular}
\end{table*}

Table~\ref{tab:opo_full} reports the relative improvement of the seven controllers. All three LLM controllers outperform the fixed-$\sigma$ baseline and the trained MetaBBO controllers on average. Qwen3.5 achieves the highest mean score of 0.511, followed by DeepSeek-V4 at 0.503 and GPT-5-mini at 0.495; each exceeds the fixed baseline of 0.417. DDQN is the strongest MetaBBO controller with an average of 0.442, but it remains below every LLM controller. PPO achieves 0.343, while OpenAI-ES ranks last at 0.211 and does not transfer effectively across the test functions. The LLM advantage is not confined to one landscape. On Rastrigin, Griewank, and Schwefel, the three LLMs achieve ranges of 0.61--0.66, 0.56--0.60, and 0.73--0.77, respectively, and outperform the strongest MetaBBO controller on each function. Ackley is the only exception. Every controller remains below 0.08 because a (1+1)-ES with one offspring per iteration cannot reliably escape its many local optima within the 1000-evaluation budget. No LLM is uniformly best. Qwen3.5 leads on Rastrigin, Schwefel, and the overall average, whereas DeepSeek-V4 leads on Ackley and Griewank. These results support the main-text finding that general LLM reasoning can match or exceed trained MetaBBO controllers on this single-parameter control task, particularly when the trained policies do not transfer across landscapes.

\begin{figure*}[ht!]
    \centering
    \includegraphics[width=0.99\textwidth]{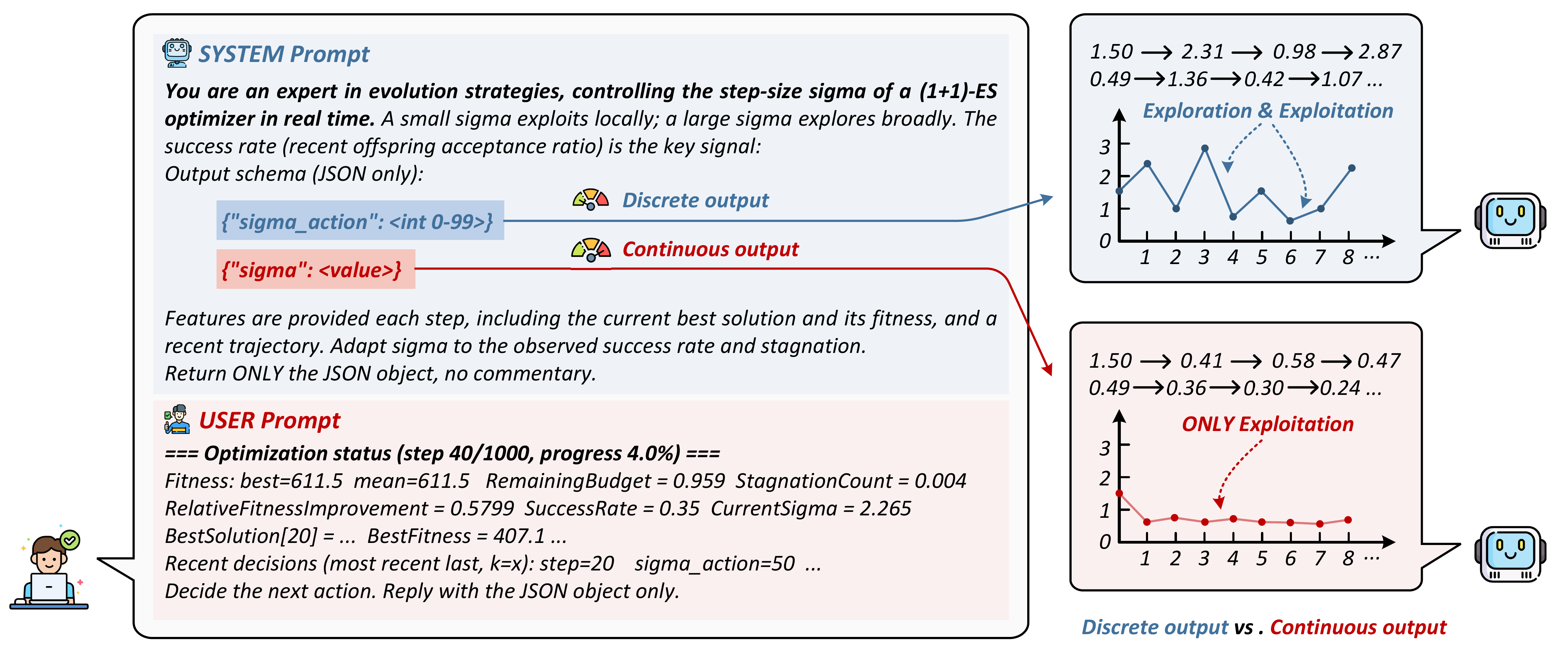}
    \caption{Representative step-size trajectories under the two output paradigms for low-level LLM assistance: direct continuous output and structured discrete output.}
    \label{fig:low_level_output}
\end{figure*}

Beyond controller performance, we examine how output representation affects LLM-based parameter control. Figure~\ref{fig:low_level_output} compares two paradigms derived from prior work \cite{custode2024investigation, zhang2025laos}. The first follows the OPRO format: it provides the full optimization trajectory and asks the LLM to output a continuous step size. The second provides a structured MetaBBO-style state and asks the LLM to select a discrete action. On Rastrigin, direct continuous output confines the step size to a narrow range and produces predominantly exploitative behavior. Structured discrete output supports more flexible transitions between exploration and exploitation. The output representation is a primary source of this difference. Continuous generation requires the model to reason directly over floating-point values, which reproduces the limitation observed in the LLM-as-optimizer experiment. Discrete actions reduce the decision space and therefore simplify control. The structured state also avoids the attention dispersion and context growth caused by an ever-expanding optimization trajectory. Recent work \cite{feng2026unleashing} further suggests that effective parameter control should move from population-level to individual-level decisions. For LLM controllers, a promising implementation is to assign parameters to buckets of batched individuals.

\section{Detailed Evaluation of High-Level Assistance \label{sec:high}}

To evaluate high-level algorithm generation in greater detail, we compare results reported by three recent comprehensive benchmarks: CO-Bench \cite{sun2026co}, HeuriGym \cite{chen2025heurigym}, and FrontierCO \cite{feng2026frontierco}. Each benchmark uses its own standardized protocol, so the comparison is intended to reveal broad regime-level trends rather than establish a single unified ranking. The benchmarks are complementary. CO-Bench provides broad coverage of classical combinatorial-optimization problems and LLM-based methods. HeuriGym focuses on low-exposure scientific and engineering problems and uses the Quality-Yield Index to distinguish solution validity from quality. FrontierCO evaluates competition-scale real-world instances that are orders of magnitude larger. Comparing the reported results across these benchmarks shows how the apparent advantage of LLM-based methods changes as problem realism and evaluation rigor increase.

\begin{figure}[ht!]
    \centering
    \includegraphics[width=0.65\linewidth]{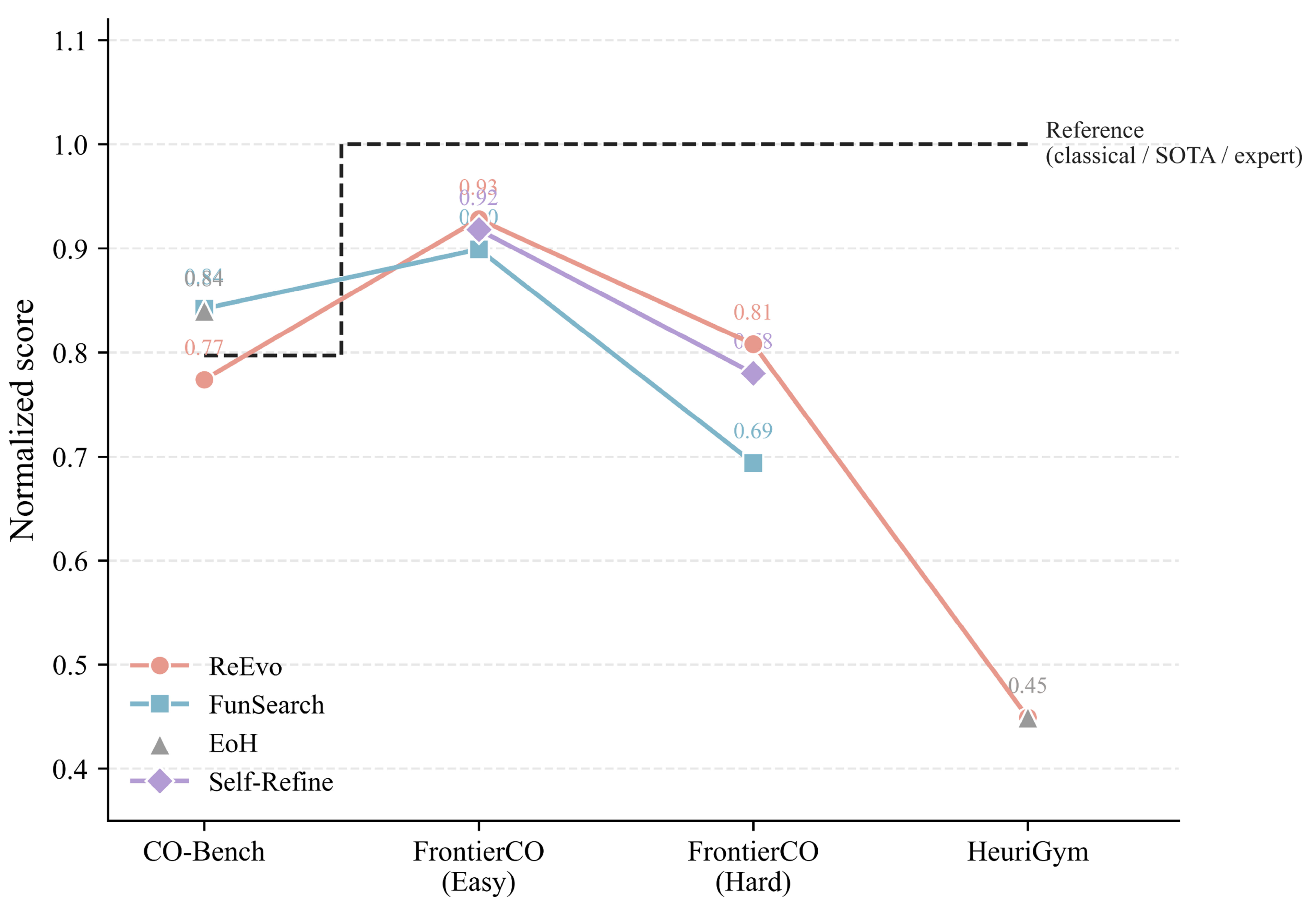}
    \caption{Performance of representative LLM-based methods across problem regimes of increasing realism. The dashed line denotes the reference baseline: a classical solver on CO-Bench and the state-of-the-art or expert baseline on the other benchmarks. Values above the line exceed the corresponding reference, while values below it do not. Scores are normalized so that 1.0 matches the best-known, state-of-the-art, or expert solution. Missing markers indicate that a method was not evaluated in that regime.}
    \label{fig:high_appendix}
\end{figure}

Figure~\ref{fig:high_appendix} shows a consistent decline as the evaluation regime becomes more realistic. On CO-Bench, FunSearch and EoH both achieve normalized scores of 0.84 and slightly exceed the classical-solver reference of 0.797, while ReEvo scores 0.77. Thus, high-level generation frameworks can rival the classical baseline on standard benchmark problems. This advantage disappears on more demanding regimes. On the FrontierCO easy set, the evaluated methods already fall below the state-of-the-art reference; FunSearch scores 0.90. On the hard set, its score falls further to 0.69. HeuriGym reveals the largest deficit: EoH and ReEvo each score 0.45, and even Gemini-2.5-Pro reaches only 0.62 against the expert reference. ReEvo is the only method evaluated in all four regimes, and its score declines from 0.77 to 0.45. Differences in reference strength explain part of this reversal. CO-Bench uses a time-limited classical solver, whereas FrontierCO and HeuriGym compare against state-of-the-art or expert performance. Exceeding the CO-Bench baseline is therefore a less demanding criterion. However, the reference alone does not explain the full decline. The methods also lose feasibility, scalability, and novelty as task realism increases.

\begin{figure}
    \centering
    \includegraphics[width=0.9\linewidth]{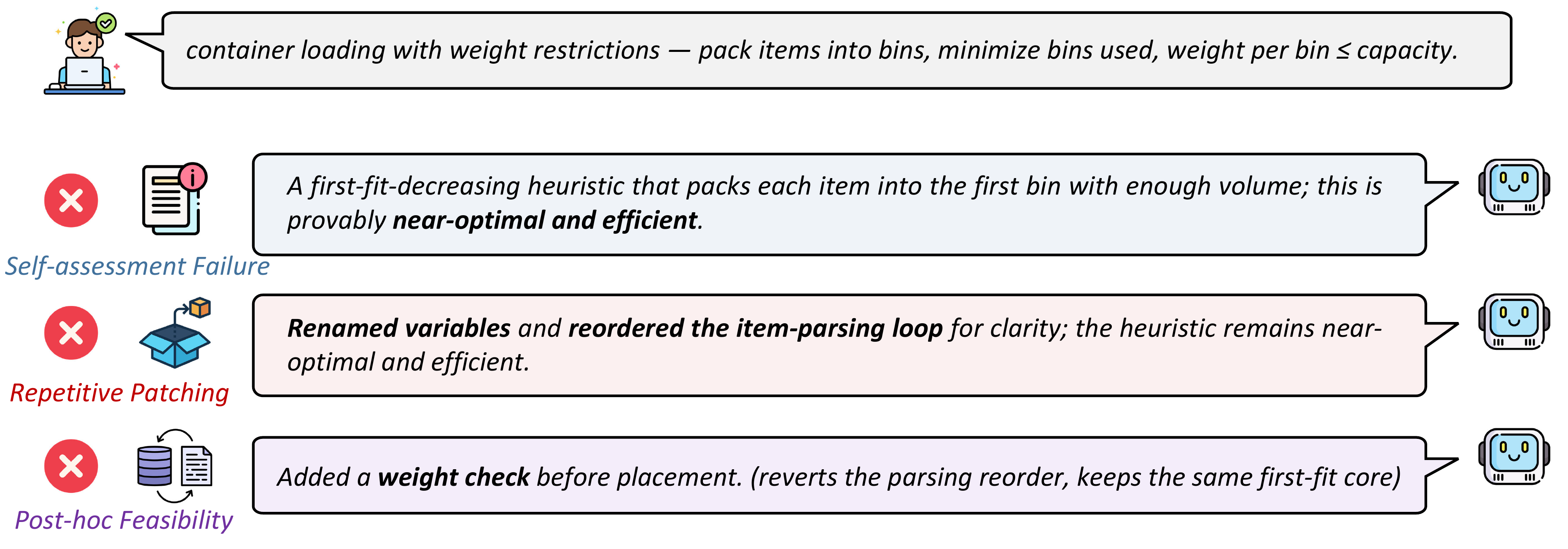}
   \caption{Representative high-level algorithm-generation failure on a container-loading task with weight constraints.}
\label{fig:highlevel_failure}
\end{figure}

Three recurring failure modes help explain this decline, and Fig.~\ref{fig:highlevel_failure} illustrates them through representative agent dialogues. First, iterative frameworks often modify the same flawed program without correcting its central defect. HeuriGym reports that broken context across iterations and poorly integrated execution feedback lead agents to rename variables or reorder unrelated loops, and agents may even return to previously failed versions. Second, agents cannot reliably assess the quality of the algorithms they generate. FrontierCO identifies this weakness as a major source of variance: an agent may describe an algorithm as near-optimal and efficient even when it times out on one routing instance, yet the same procedure may outperform the state of the art on another. The agent cannot distinguish these outcomes in advance. Third, feasibility and efficiency are often checked only after generation. Some heuristics violate constraints during construction, while others use local-search depths with exponential cost and therefore time out on large instances. As a result, even the strongest agents produce valid solutions less consistently than the classical solver. These failures identify the central research frontier: reliable self-assessment, cumulative use of feedback, and constraint-aware reasoning.

\end{document}